%% file: main.tex
\documentclass[10pt,twocolumn,letterpaper]{article}
\usepackage{hyperref}
\pdfoutput=1 
\usepackage{cvpr}
\usepackage{times}
\usepackage{epsfig}
\usepackage{graphicx}
\usepackage{amsmath}
\usepackage{amssymb}
\usepackage{bm}
\usepackage{subfigure}
\usepackage{epstopdf}
\usepackage{mathrsfs}
\usepackage[font=small,skip=-5pt]{caption}
\usepackage{multirow,array}

\usepackage{algorithm} %format of the algorithm
\usepackage{algorithmic} %format of the algorithm

\usepackage{array}
\usepackage{tabularx}
\cvprfinalcopy % *** Uncomment this line for the final submission

\def\cvprPaperID{****} % *** Enter the CVPR Paper ID here

\ifcvprfinal\fi
\begin{document}

%%%%%%%%% TITLE
\title{Baseline Desensitizing In Translation Averaging}

\author{Bingbing Zhuang,~~Loong-Fah Cheong,~~Gim Hee Lee\\
	National University of Singapore\\
	{\tt\small zhuang.bingbing@u.nus.edu, \{eleclf,gimhee.lee\}@nus.edu.sg}
}

\maketitle
\thispagestyle{empty}
%%%%%%%%% ABSTRACT
\begin{abstract}
	\vspace{-0.2cm}
  Many existing translation averaging algorithms are either
 sensitive to disparate camera baselines and have to rely
 on extensive preprocessing to improve the observed
 Epipolar Geometry graph, or if they are robust against
 disparate camera baselines, require complicated optimization
 to minimize the highly nonlinear angular error objective.
 In this paper, we carefully design a simple yet effective bilinear
 objective function, introducing a variable to perform
 the requisite normalization. The objective function enjoys the
 baseline-insensitive property of the angular error and
 yet is amenable to simple and efficient optimization by block coordinate
 descent, with good empirical performance. A rotation-assisted Iterative 
 Reweighted Least Squares scheme is further put forth to help deal with outliers.
 We also contribute towards a better understanding of the behavior
 of two recent convex algorithms, LUD \cite{ozyesil2015robust} and Shapefit/kick
 \cite{goldstein2016shapefit}, clarifying the underlying subtle difference that
 leads to the performance gap. Finally, we demonstrate that our algorithm
 achieves overall superior accuracies in benchmark
 dataset compared to state-of-the-art methods, and is also
 several times faster.
\end{abstract}

%%%%%%%%% BODY TEXT
	\vspace{-0.2cm}
\input{Section/sec_introduction}
\input{Section/sec_relatedworks}

%\input{Section/sec_background}
\input{Section/sec_method}

\input{Section/sec_experiment}
\input{Section/sec_conclusion}

{\small
\bibliographystyle{ieee}
%\bibliography{egbib}

\input{main.bbl}
}

\end{document}

%% file: Section/sec_introduction.tex
\section{Introduction}
%Structure-from-Motion (SfM) has been one of the most successful branch in computer vision community over the last decades. Especially, the popularity of social networks, e.g. Flickr, has given rise to modern large scale SfM systems \cite{snavely2006photo,wu2011visualsfm,theia-manual,schoenberger2016sfm} from unordered image collections from Internet . 
Modern large-scale Structure-from-Motion (SfM) systems have enjoyed widespread success in many applications \cite{snavely2006photo,wu2011visualsfm,theia-manual,schoenberger2016sfm,moulon2016openmvg}.
Earlier methods often adopt an incremental method by adding the cameras one by one sequentially as the size of the model grows up. As a consequence, the quality of the result heavily depends on the order in which the cameras are added, and the accumulated error often leads to significant drift as the size of the model increases. Therefore, frequent intermediate bundle adjustments (BA) \cite{triggs1999bundle} have to be applied to obtain stable result, which unfortunately increases the computational cost substantially. Given such disadvantages, the global SfM method emerges as a serious alternative. Unlike the incremental method, the global method attempts to determine the absolute poses for all the cameras simultaneously from all the observed pairwise Epipolar Geometry (EG) \cite{hartley2003multiple}. Such a holistic approach spreads the error as uniformly as possible to the whole model, avoiding the problem of error accumulation and drift.
%This shares the similar elegance in factorization-based SfM \cite{tomasi1992shape}.
Thus, BA needs to be run only once as a final refinement, leading to a more efficient system.

Formally, a global SfM algorithm takes as input a view graph $G{=}(V,E)$, where each node $V_i$ in $V$ and edge $E_{ij}$ in $E$ represent respectively a camera and relative pose $(\bm{R}_{ij},\bm{t}_{ij})$ between the camera pair $i$ and $j$ whose fields of view overlap. It aims to find the absolute rotation $\bm{R}_{i}$ (a.k.a. rotation averaging) and location $\bm{t}_{i}$ (a.k.a. translation averaging) for each camera (up to a gauge freedom), such that the observed pairwise relative poses are best explained. In the noiseless case, the following two equations hold:
\vspace{-0.08cm}
\begin{equation}
\bm{R}_i^T\bm{R}_j = \bm{R}_{ij}, \quad \frac{\bm{t}_j-\bm{t}_i}{||\bm{t}_j-\bm{t}_i||_2} = \bm{R}_i\bm{t}_{ij}.
\end{equation}
Typically, rotation averaging is performed before translation averaging. In this paper, we follow this practice and shall focus on the second equation to perform translation averaging. Note that rotation is assumed to have been solved, and henceforth, for brevity, we shall denote $\bm{R}_i\bm{t}_{ij}$ as $\bm{v}_{ij}$.

Translation averaging is recognized as a hard task.
%In fact, the now-common practice of separating rotation averaging and translation averaging into two sequential steps, which we also follow in our framework, is due to the above difficulty gap, since such separation prevents from adjusting rotation solution to complement the error from translation.
One of the reasons is that the input relative translation estimate is sensitive to small camera baselines \cite{enqvist2011non}. More importantly, EG only encodes the relative direction between cameras without any magnitude information. This causes a remove of the measurement space (directions between pairwise cameras) from the solution space (camera locations); this gap complicates the task much more, posing a significant challenge for the objective function design. The 
%statistically superior objective 
geometrically more meaningful objective
would be to minimize the angular error between unit direction vectors \cite{sim2006recovering,wilson2014robust}, but this leads to highly nonlinear functions that require complicated optimization. Instead, many recent methods simply ignore the normalization terms required for obtaining unit vectors, thereby yielding various forms of quasi-Euclidean distance terms in the objectives \cite{moulon2013global,ozyesil2015robust,goldstein2016shapefit}. Often, such expediency allows the problem to be formulated as a convex optimization problem.
%or even permits exact recovery guarantee under some conditions in theory \cite{goldstein2016shapefit}. 
However, the serious qualification of such magnitude-based objective functions is that 
%the solution suffers from bias when the camera baselines are disparate in lengths.
they suffer from unbalanced weighting on each individual term when the camera baselines are disparate in lengths, which may lead to biased solutions.
These methods usually employ extensive preprocessing (e.g. outlier filtering) of the view graph to obtain better relative translations as input. This relieves but does not resolve the issue fundamentally and the accuracy is still limited in practice.

The contributions in this paper are threefold. (1) We show that by carefully designing the objective function, the numerical sensitivity of the solution to different camera baseline lengths can be readily removed. Specifically, we propose a return to the geometrically more meaningful, angular-error based objective function, putting forth a simple yet accurate Bilinear Angle-based Translation Averaging (BATA) framework. The key idea is to introduce a variable that performs the requisite normalization for a baseline-insensitive angular error term. This splits the original problem into easier subproblems, which can be easily optimized  by block coordinate descent; empirically, the algorithm converges fast and yields superior performance.
(2) To deal with outlier EG,  we put forth a rotation-assisted Iterative Reweighted Least Squares (IRLS) scheme that leverages on the stable solution from rotation averaging as an extra source of information to determine the reliability of each observation in the view graph. (3) Our objective formulation also lends perspective to the behavior of various algorithms with a magnitude-based objective function \cite{moulon2013global,ozyesil2015robust,goldstein2016shapefit}. We reveal that the subtle difference in the scale ambiguity removal strategy can nevertheless lead to rather different performance in such algorithms.
Specifically, we build the equivalence between Shapefit/kick \cite{goldstein2016shapefit} and a slightly revised version of LUD \cite{ozyesil2015robust}, which allows us to trace the difference between  Shapefit/kick and LUD to the scale ambiguity removal constraint. We then demonstrate that a weaker lower-bound constraint can cause a squashing effect on the overall shape of the recovered camera locations, especially under the presence of disparate baselines; conversely, a stronger constraint would help desensitize the effect of unbalanced baselines.

We demonstrate the utility of the proposed framework by extensive experiments on both synthetic and real data. In particular,  we obtain superior performance on the benchmark 1DSfM dataset \cite{wilson2014robust} both in terms of accuracy and efficiency compared to state-of-the-art methods. The code will be made publicly available.

%% file: Section/sec_relatedworks.tex
\section{Related Work}
\noindent\textbf{Rotation Averaging.}
Many methods exist for this task \cite{hartley2013rotation,chatterjee2013efficient,martinec2007robust,arrigoni2014robust,fredriksson2012simultaneous,govindu2001combining}; we refer readers to \cite{tron2016survey} for a survey.

\noindent\textbf{View Graph Preprocessing.} Some methods \cite{zach2010disambiguating,wilson2014robust} utilize loop consistency to remove outlier EG in the view graph. 
%Although effective, such filtering schemes often have difficulty in determining how many edges to filter out; removing more edges increases the risk of losing the parallel rigidity of the graph, a condition needed for the camera pose to have a unique solution \cite{ozyesil2015robust}. 
Some other works attempt to refine the whole view graph using loop consistency \cite{sweeney2015optimizing,shen2016graph} or low-rank constraint \cite{Sengupta_2017_CVPR}. A robust re-estimation of the pairwise translation after the recovery of absolute rotation is proposed in \cite{ozyesil2015robust}.

\noindent\textbf{Translation Averaging.} The pioneering work by Govindu \cite{govindu2001combining} proposes to minimize the cross product between the relative camera location $\bm{t}_j{-}\bm{t}_i$ and the observed direction $\bm{v}_{ij}$. An ad-hoc iterative reweighting scheme is adopted to reduce the bias from different baseline lengths. As reported in \cite{wilson2014robust}, this generates poor accuracy in challenging dataset. 
%\cite{arie2012global} also uses the cross product term but without baseline normalization (akin to the linear epipolar constraint); such method, however, suffers from the well-known bias from minimizing an algebraic error that is not geometrically meaningful.
Some methods aim to minimize the relaxation of the endpoint distance $\big\lVert\bm{t}_j{-}\bm{t}_i{-}\lVert\bm{t}_j{-}\bm{t}_i\rVert_2\bm{v}_{ij}\big\rVert_2^2$ or its variants.
For example, Moulon et al. \cite{moulon2013global} propose to minimize a relaxed version using the $L_\infty$ norm. A similar penalty is utilized in \cite{ozyesil2015robust} but with a least unsquared deviations (LUD) form to be more robust. Goldstein et al. \cite{goldstein2016shapefit} propose a Shapefit/kick scheme based on the alternating direction method of multipliers (ADMM) to minimize the magnitude of the projection of $\bm{t}_j{-}\bm{t}_i$ on the orthogonal complement of $\bm{v}_{ij}$. Despite its convex formulation, works such as \cite{moulon2013global,ozyesil2015robust,goldstein2016shapefit} suffer from bias due to the unnormalized camera baseline magnitude in their objectives, and often have to resort to extensive preprocessing strategies reviewed in the preceding paragraph to take more accurate EG view graph as input.

Works that minimize the angular residual between $\bm{t}_j{-}\bm{t}_i$ and $\bm{v}_{ij}$, denoted as $\theta_{ij}$, are relatively rare in the literature.  One of the representative works is that of Sim and Hartley \cite{sim2006recovering}.  They show that minimizing the maximal absolute value of $\tan \theta_{ij}$ from all observations, i.e. $L_\infty$ norm, can be reformulated into a quasi-convex problem and a globally optimal solution can be found by solving a sequence of Second Order Cone Programming (SOCP) feasibility problems. However, it is well known that $L_\infty$ is sensitive to outliers, and solving multiple SOCP problems restricts their method to medium-size problems. Wilson et al. \cite{wilson2014robust} present another attempt by minimizing the residual of $\sin \theta_{ij}/2$. The trust-region method Levenberg-Marquard is applied to optimize the resultant highly nonlinear function. In our work, we present another objective function along this line of approach; it minimizes the residual of $\sin\theta_{ij}$ in essence, is easily optimizable and yet achieves superior performance.

Other heuristics have been proposed. These include coplanar constraint on triple cameras \cite{jiang2013global}, reprojection error \cite{kahl2005multiple,martinec2007robust},
% registering pairwise reconstruction to obtain a global one \cite{sinha2010multi}, 
reducing the problem into similarity averaging by local depth estimation \cite{cui2015global}, and others \cite{sinha2010multi,crandall2011discrete,tron2014distributed,olsson2011stable,arie2012global}. 
We note that existing methods often include scene points to assist translation averaging and/or involve careful outlier filtering step. 
%In this paper, we present a more concise framework and demonstrate the possibility of achieving good accuracy in practice even if we directly process the raw view graph. Such framework is more amenable to efficient processing due to much smaller number of variables. 
In this paper, we demonstrate the possibility of achieving good accuracy in practice even if we directly process the raw view graph. Such a concise framework is more amenable to efficient processing. 

%Such methods often need careful outlier filtering step and/or are computationally expensive. We also note that existing frameworks fall into either reconstructing camera positions only (e.g. \cite{jiang2013global,sim2006recovering,moulon2013global} ) or involving the reconstruction of scene points (e.g. \cite{wilson2014robust,cui2015global,olsson2011stable}). The former is more amenable to efficient processing due to much smaller number of variables; our framework belongs to this category. 

%% file: Section/sec_method.tex
\vspace{0.1cm}
\section{Method}
%\begin{figure}
%	\setlength{\abovecaptionskip}{-0.45cm}
%	\setlength{\belowcaptionskip}{-0.45cm}
%	\begin{center}
%		\includegraphics[width=1\linewidth, trim = 65mm 105mm 45mm 20mm, clip]{Figure/geometricmeaning.pdf}
%	\end{center}
%	\caption{Geometric interpretation of the residual penalized by different objective functions. (a) $\theta_{ij}$. (b) $||\bm{t}_j-\bm{t}_i-||\bm{t}_j-\bm{t}_i||_2\bm{v}_{ij} ||_2^2$. (c) 1DSfM.
%		(d) Shapefit/kick and RevisedLUD. (e) Ours. Refer to the text for more details.}
%	\label{fig:geometricmeaning}
%\end{figure}

\begin{figure}
	\setlength{\abovecaptionskip}{-0.2cm}
	\setlength{\belowcaptionskip}{-0.45cm}
	\begin{center}
		\includegraphics[width=1\linewidth, trim = 70mm 100mm 22mm 40mm, clip]{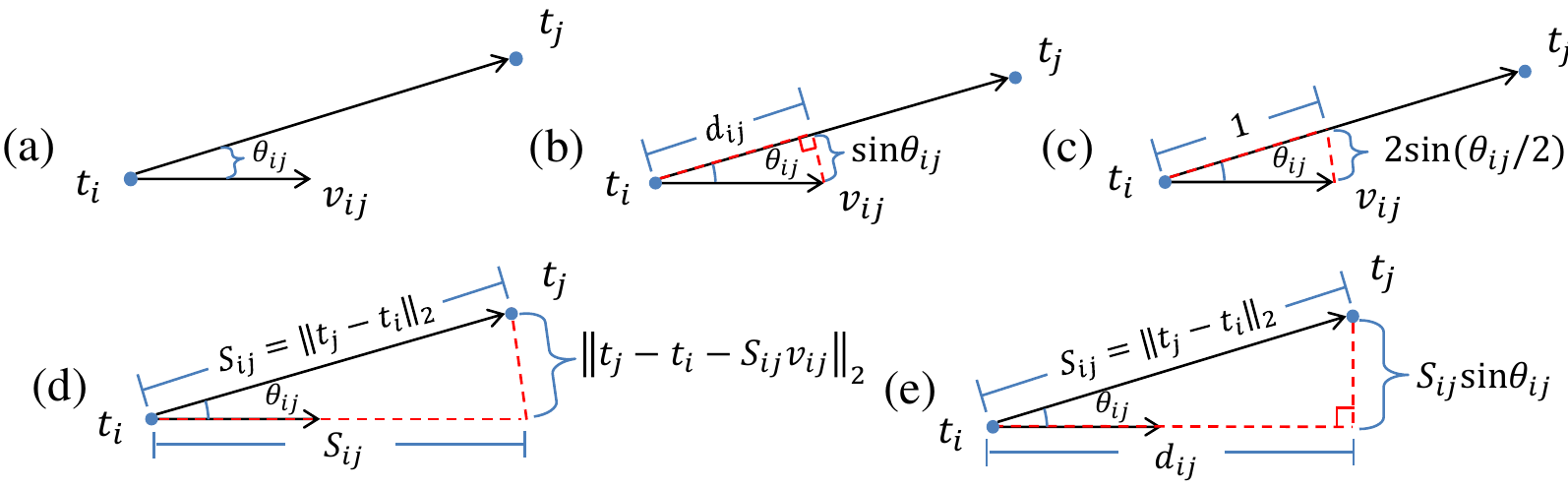}
	\end{center}
	\caption{Geometric interpretation of the residuals penalized by different objective functions. (a) $\theta_{ij}$. (b) BATA. (c) 1DSfM.
		(d)  $\big\lVert\bm{t}_j{-}\bm{t}_i{-}\lVert\bm{t}_j{-}\bm{t}_i\rVert_2\bm{v}_{ij} \big\rVert_2$.  (e) Shapefit/kick and RevisedLUD. Refer to the text for more details.}
	\label{fig:geometricmeaning}
\end{figure}

\vspace{0.1cm}
\subsection{Bilinear Angle-based Translation Averaging}
\label{sec:objfunc}

\begin{figure}
	\setlength{\abovecaptionskip}{-0.2cm}
	\setlength{\belowcaptionskip}{-0.4cm}
	\begin{center}
		\includegraphics[width=1\linewidth, trim = 60mm 45mm 48mm 5mm, clip]{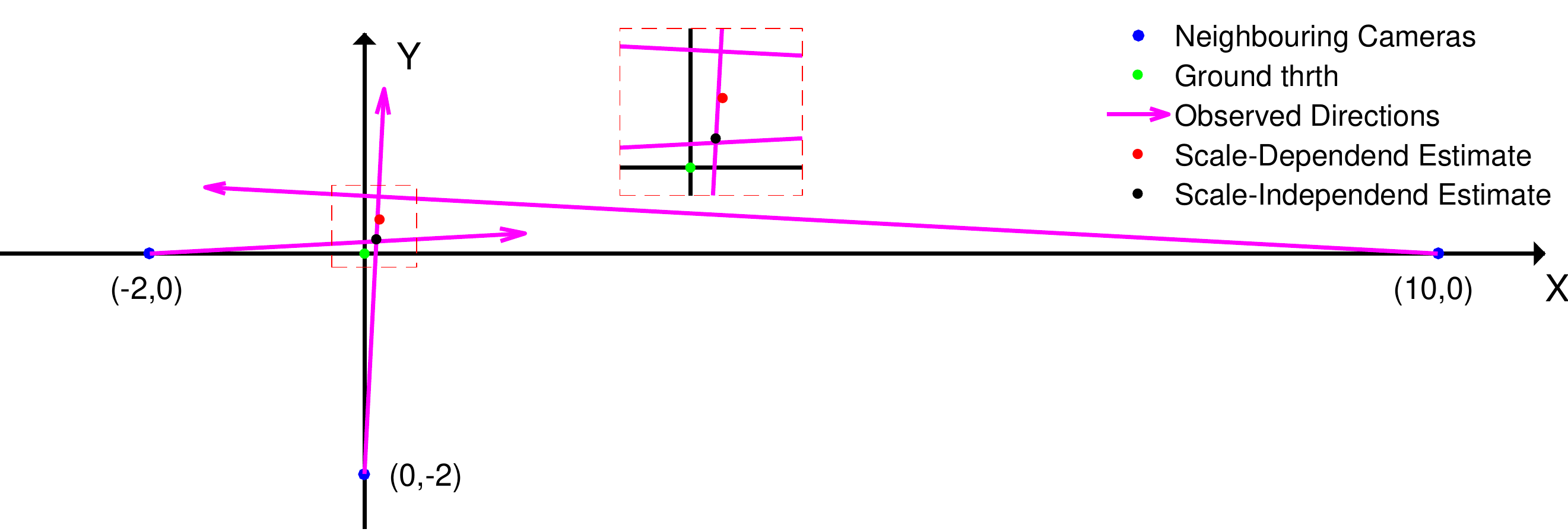}
	\end{center}
	\caption{A toy example showing the sensitivity of different objectives to disparate baselines. See text for more explanations.}
	\label{fig:toyexample}
\end{figure}

Instead of penalizing the angular deviation $\theta_{ij}$ between \newpage
\hspace{-0.5cm} $\bm{t}_j{-}\bm{t}_i$  and the observed $\bm{v}_{ij}$ (illustrated in Fig.~\ref{fig:geometricmeaning}(a)) directly, many existing algorithms adopt an objective function of the form $\sum_{ij \in E}\big\lVert\bm{t}_j{-}\bm{t}_i{-}\lVert\bm{t}_j{-}\bm{t}_i\rVert_2\bm{v}_{ij} \big\rVert_2^2$ (Fig.~\ref{fig:geometricmeaning}(d)) or its variants. Note that this objective function is not normalized by the vector magnitude $\lVert\bm{t}_j{-}\bm{t}_i\rVert_2$, and is thus plagued by numerical difficulties when there is large variation in the magnitudes of $\bm{t}_j{-}\bm{t}_i$. In particular, though this objective will yield the true solution in the absence of noise, it is not necessarily statistically optimal under noisy condition. Let us illustrate this point via a toy example in the 2D plane. Referring to Fig.~\ref{fig:toyexample}, suppose we know the ground-truth locations of three neighboring cameras to be at (-1, 0), (0, -1) and (5, 0) and would like to localize the fourth camera, with ground truth at (0, 0), according to its observed pairwise directions with respect to the three neighboring cameras. Due to noise, suppose all these observed directions deviate from their true direction by $3^\circ$. We use the red dot to denote the best location (found by exhaustive search on 2D grid) that minimizes the preceding magnitude-based residual. We also use the black star to denote the best location that minimizes the squared angular deviation $\sum_{ij \in E}\theta_{ij}^2$. Clearly, the solution in the former case is much worse off compared to that in the latter. In the former case, the objective function is essentially trying to determine the intersection point of the three direction vectors (in some least squares sense). This process is highly susceptible to errors when one or more cameras are far away. It follows that using such a magnitude-based objective function for the translation averaging problem would also experience similar sensitivity issue when there are disparate camera baseline distances.

In view of the foregoing discussion, we propose the following angle-based objective function instead:

\vspace{-0.5cm}
\begin{align}\label{eq:optim_formu}
\min_{\substack{\bm{t}_i,i\in V,\\d_{ij},ij\in E}} & \quad \sum_{ij\in E}\rho(||(\bm{t}_j-\bm{t}_i){d_{ij}}-\bm{v}_{ij}||_2),\\
s.t. \quad & \sum_{i\in V} \bm{t}_{i} = 0, \sum_{ij\in E} \langle \bm{t}_j-\bm{t}_i,\bm{v}_{ij} \rangle = 1, \notag \\
& d_{ij} \geq 0, \forall ij\in E, \notag
\end{align}

\noindent where $\rho(\cdot)$ stands for a robust M-estimator function to be discussed in the next section. $d_{ij}$ is a non-negative variable. The first two constraints on $\bm{t}$ are to remove the inherent positional and scale ambiguity. We now show that the optimal solution of this problem essentially minimizes an angular error $\sum_{ij\in{E}}\rho(h(\theta_{ij}))$, where
\vspace{-0.4cm}
\begin{equation}\label{eq:sin_theta}
h(\theta_{ij})= \left\{
\begin{aligned}
&\sin \theta_{ij}, &\theta_{ij} \leq 90^\circ; \\
& 1,   & \theta_{ij} > 90^\circ.
\end{aligned}
\right.
\end{equation}

\vspace{-0.4cm}
\noindent First, note that for any given candidate solution $\hat{\bm{t}}$, each $d_{ij}$ serves to scale the relative location vector $\hat{\bm{t}}_j{-}\hat{\bm{t}}_i$ such that the Euclidean distance between the endpoint of $(\hat{\bm{t}}_j{-}\hat{\bm{t}}_i){d_{ij}}$ and the unit vector $\bm{v}_{ij}$ is minimized. It follows that if the angle between $\hat{\bm{t}}_j{-}\hat{\bm{t}}_i$ and $\bm{v}_{ij}$ is less than $90^\circ$, the optimal value ${d}_{ij}$ would be such that $(\hat{\bm{t}}_j{-}\hat{\bm{t}}_i){{d}_{ij}}$ equals to the projection of $\bm{v}_{ij}$ onto the direction along $\hat{\bm{t}}_j{-}\hat{\bm{t}}_i$ and the penalty amounts to $\sin\theta_{ij}$ (Fig.~\ref{fig:geometricmeaning}(b)). The constraint $d_{ij}>0$ prevents $d_{ij}$ from overcompensating when $\theta_{ij}>90^\circ$, otherwise the objective would decrease when $\theta_{ij}$ increases from $90^\circ$ to $180^\circ$. With this constraint, the optimal ${d}_{ij}$ when $\theta_{ij}>90^\circ$ would be 0 and the penalty would be 1.

As an alternative viewpoint, we could have regarded our objective function as a functional lifting and relaxation of the formulation  in 1DSfM \cite{wilson2014robust} which is an unconstrained minimization of the objective function $\sum_{ij \in E} \rho(\big\lVert (\bm{t}_j{-}\bm{t}_i)/\lVert\bm{t}_j{-}\bm{t}_i\lVert_2{-}\bm{v}_{ij}\big\rVert_2)$.  The scale factor $1/\lVert\bm{t}_j{-}\bm{t}_i\rVert_2$ is replaced with the variable $d_{ij}$ together with a relaxation of the constraint from $d_{ij}{=}1/\lVert\bm{t}_j{-}\bm{t}_i\rVert_2$ to $d_{ij}{>}0$. We note that since the scale factor in 1DSfM always normalizes the vector ${\bm{t}}_j{-}{\bm{t}}_i$ to a unit vector, its objective amounts to the penalty term $2\sin(\theta_{ij}/2)$ (Fig.~\ref{fig:geometricmeaning}(c)), which bears a close resemblance to the $\sin \theta_{ij}$  established in (\ref{eq:sin_theta}). It is clear that without any prior knowledge on the noise distribution, there is no reason for one to claim superiority over the other. Thus, while one can regard our objective function as a relaxation of that of 1DSfM, one should not see the relaxed version as a poorer cousin of the two, since there is nothing sacrosanct about $2\sin(\theta_{ij}/2)$ over $\sin\theta_{ij}$ in terms of its geometrical meaning. We also note that the relaxation reduces the original highly nonlinear term into a bilinear one that permits simple alternating optimization; compared to the solution of 1DSfM from Ceres \cite{ceres-solver}, we empirically observe %that BATA can recover those sparsely connected cameras more reliably in real challenging data 
that BATA can generally  recover the camera locations more reliably in real Internet photo collections, especially for those challenging sparsely connected cameras. 
%and might have smoothed the objective function and reduced its chance of getting stuck in some bad local minimum. 
%In fact, empirically, we observe superior accuracy from BATA in both the synthetic data with the observations contaminated by Gaussian noise and in the real-world 1DSfM dataset \cite{wilson2014robust}. More evaluations and analysis will follow in the experimental section.
Also note that our penalty term in (\ref{eq:sin_theta}) levels off after $\theta_{ij}{>}90^\circ$, and this might bestow greater robustness to our formulation.

\subsection{Robust Rotation-Assisted IRLS}
\label{sec:IRLS}
As the estimated EG view graph often contains gross outliers, we thus embed the least squares objective into a M-estimator $\rho(\cdot)$. Iterative Reweighted Least Squares scheme is often used to optimize such objective, whereby a weighted least squares problem is solved in each iteration. The weight function, denoted as $\phi(\cdot)$ here, returns a value proportional to the goodness of fit of an observation $ij$, evaluated at the last iteration. Note that the specific form of $\phi(\cdot)$ depends on the M-estimator function $\rho(\cdot)$ being used, e.g. for Cauchy $\rho(\varepsilon ){=}log(1+\varepsilon^2/\alpha^2)$ and $\phi(\varepsilon){=}\alpha^2/(\alpha^2+\varepsilon^2)$, where $\varepsilon$ denotes the residual for each observation and $\alpha$ is the loss width. 
%Given that rotation averaging can often be computed more reliably \cite{hartley2013rotation,chatterjee2013efficient}, 
Since it is well known that rotation averaging can often be computed more reliably \cite{hartley2013rotation,chatterjee2013efficient}, 
a natural idea is to leverage its result to assist the reliability assessment or weighting for each observed EG.
We thus use for this purpose the following residual $\varepsilon = (\lVert(\bm{t}_j{-}\bm{t}_i)d_{ij}{-}\bm{v}_{ij}\rVert_2^2+\beta\lVert\bm{R}_i^T\bm{R}_j{-}\bm{R}_{ij} \rVert_2^2)^{1/2}$,
whereby the goodness of fit of the rotation estimate also contributes to the weighting process. Here, $\beta$ is a predefined weighting factor (set as 1 for all the experiments). 
%This strategy turns out to be generally effective and allows BATA to achieve good accuracy even from random initialization.
It turns out that this strategy can generally improve the accuracy and speed up the convergence of BATA. 
For each IRLS iteration, we use Block Coordinate Descent (BCD) to optimize $\bm{t}$ and $d$, as summarized in Algo.~\ref{algo:irlsbcd}. 

\vspace{-0.35cm}
\begin{algorithm} \footnotesize
	\caption{IRLS-BCD solver}
	\label{algo:irlsbcd}
	\begin{algorithmic}[1] 	
		\REQUIRE  View Graph $G = (V,E)$, Rotation Averaging Result.
		\ENSURE Camera Locations $\bm{t}_i, \forall i \in V$.
		\STATE Initialize $\bm{t}_i, \forall i \in V, W_{ij}, \forall ij\in E$; Set $n = 0$; 
			\WHILE{$n <$ IRLSIter AND not converged }      					
				\STATE $m = 0$;
				\WHILE{ $m <$ BCDIter} 
				\STATE \textbf{Update} $d_{ij}:$ $d_{ij} = max( \frac{\left\langle\bm{t}_j-\bm{t}_i,\bm{v}_{ij}\right\rangle}{||\bm{t}_j-\bm{t}_i||_2^2},0)$;
				\STATE \textbf{Update} $\bm{t}_{i}:$ Solve a sparse, weighted, constrained linear least squares system of equations collected from (\ref{eq:optim_formu}) by Cholesky decomposition (see \emph{supp. material} for more details); 							
				\STATE $m = m + 1$;
				\ENDWHILE 
				\STATE  \textbf{Update} $W_{ij}$: \footnotesize{$W_{ij} = \phi(\varepsilon)$, where \\
				\hspace{1.6cm} $\varepsilon=(\lVert(\bm{t}_j{-}\bm{t}_i)d_{ij}{-}\bm{v}_{ij}\rVert_2^2{+}
				\beta\lVert\bm{R}_i^T\bm{R}_j{-}\bm{R}_{ij} \rVert_2^2)^\frac{1}{2}$};				
				\vspace{-0.4cm}
				\STATE $n = n + 1$; 	
		\ENDWHILE 
		% repeat-until loop 				
	\end{algorithmic} 
\end{algorithm}

\iffalse
For a comprehensive comparison, we would also try an heuristic reweighing attempt similar to the scheme in \cite{govindu2001combining} by simply setting $d_{ij} = 1/||\bm{t}_j - \bm{t}_i||$ in each iteration of alternating optimization. As can be seen later, this heuristic scheme gives poorer accuracy in practice. Since our auxiliary variable scheme does not make the problem significantly more complex, it is always recommended.
\fi

\vspace{-0.35cm}
\subsection{Why does Shapefit/kick outperform LUD?} 
\label{sec:whatLUD}
Using the same geometric analysis, we are now ready to clarify why Shapefit/kick \cite{goldstein2016shapefit} outperforms LUD \cite{ozyesil2015robust} (if run on the same problem instances), as reported in \cite{goldstein2016shapefit} and verified by our experiments. For ease of discussion, we present their respective formulations below. 

\vspace{0.06cm}
\noindent \textbf{LUD:}
\vspace{-0.5cm}
%\begin{subequations}
	\begin{align}	\label{eq:LUD}
\hspace{-0.3cm}	\min_{\substack{\bm{t}_i,i\in V;\\d_{ij},ij\in E}} & \quad \sum_{ij\in E}||\bm{t}_j-\bm{t}_i-d_{ij}\bm{v}_{ij}||_2,\\
	s.t. \quad & \sum_{i\in V} \bm{t}_{i} = 0; d_{ij} \geq c, \forall ij\in E, \notag
	\end{align}

%\end{subequations}
\vspace{-0.3cm}
\noindent \hspace{-0.1cm} where $d_{ij}$ is deemed as a relaxation of $\lVert\bm{t}_j-\bm{t}_i\rVert_2$.

\noindent\textbf{Shapefit/kick:}
\vspace{-0.4cm}
%\begin{subequations}
	\begin{align}	\label{eq:shapefit/kick}
\quad \quad	\min_{\bm{t}_i,i\in V} & \sum_{ij\in E}\lVert P_{\bm{v}_{ij}^\perp}(\bm{t}_j-\bm{t}_i)\rVert_2,\\
	s.t. \quad & \sum_{i\in V} \bm{t}_{i} = 0, \sum_{ij\in E} \langle\bm{t}_j-\bm{t}_i,\bm{v}_{ij}\rangle = 1, \notag
	\end{align}

%\end{subequations}
\vspace{-0.4cm}

\noindent where $P_{\bm{v}_{ij}^\perp}$ denotes the projection onto the orthogonal complement of the span of $\bm{v}_{ij}$.

To tease out the connection between these formulations, 
%we replace the constraint used for removing scale ambiguity and preventing all cameras collapsing to a single point (under such case the penalty cost vanishes) in LUD, i.e. $d_{ij}\geq c$, with the one used in Shapefit/kick, i.e. $\sum_{ij\in E}\langle\bm{t}_j-\bm{t}_i,\bm{v}_{ij}\rangle = 1$. 
we replace the LUD's constraint $d_{ij}\geq c$, which is for removing scale ambiguity and preventing all cameras from collapsing to a single point (under such case the penalty cost vanishes), with the one used in Shapefit/kick, i.e. $\sum_{ij\in E}\langle\bm{t}_j{-}\bm{t}_i,\bm{v}_{ij}\rangle {=} 1$. 
We claim that the resultant optimization problem, denoted as RevisedLUD, has exactly the same optimal solution as that of Shapefit/kick. To verify this, we note that removing the constraint $d_{ij}\geq c$  reduces $d_{ij}$ to a completely free variable. Similar to the analysis for our formulation (\ref{eq:optim_formu}), for a set of estimates $\hat{\bm{t}}_i$, $\forall i \in V$, the optimal ${d}_{ij}$ would be such that ${d}_{ij}\bm{v}_{ij}$ equals to the projection of $\hat{\bm{t}}_j{-}\hat{\bm{t}}_i$ onto the direction along $\bm{v}_{ij}$ . It is immediately clear that the residual being minimized in RevisedLUD is $\lVert\bm{t}_j{-}\bm{t}_i\rVert_2\sin\theta_{ij}$  (Fig.~\ref{fig:geometricmeaning}(e)) for any $\theta_{ij}$. It is also clear that $\lVert P_{\bm{v}_{ij}^\perp}(\bm{t}_j{-}\bm{t}_i)\rVert_2$ coincides with $\lVert\bm{t}_j{-}\bm{t}_i\rVert_2\sin\theta_{ij}$ and this establishes our claim. We note here that RevisedLUD can also be assisted by rotation when optimized with IRLS in Algo.~\ref{algo:irlsbcd} with small changes (e.g. step 4 becomes $d_{ij}{=}\left\langle\bm{t}_j{-}\bm{t}_i,\bm{v}_{ij}\right\rangle$). We will have occasion to use this later when a convex initialization for BATA is called for.

Given the above equivalence, we can restrict our attention to the sole difference between the two algorithms, namely in the constraints discussed above, with a view to elucidating the impact that different formulations of these constraints have on performance. First observe that while the constraint used in RevisedLUD fixes the overall scale to a constant value, the one in LUD, i.e. $d_{ij}\geq c$, only imposes a lower bound on the scale. We note that the latter is a weaker constraint, in the sense that while it prevents the collapsing of cameras all the way to a single point, it still allows a partial shrinking to occur, i.e. shrinking without respecting the overall global shape.
%That is, in the presence of noise, those camera pairs with larger baseline often cause larger residual in $||\bm{t}_j-\bm{t}_i-d_{ij}\bm{v}_{ij}||_2$, where $d_{ij}>c$ is also more likely to happen. Since $d_{ij}\geq c$ does not impose explicit constraint on these edges, it gains more to shrink the pairwise distance in these pairs than those edges where $d_{ij}$ is bounded on $c$. 
To see the difference more explicitly, suppose we feed the optimal solution $\bm{t}^S$ obtained from the more strongly constrained (\ref{eq:shapefit/kick}) into (\ref{eq:LUD}). The solution space in (\ref{eq:LUD}) can be conceptually distinguished into two optimization regimes 1 and 2. Regime 1 admits solution of the form $\gamma\bm{t}^S$ where $\gamma$ is a scale to be optimized together with $d_{ij}$'s. Clearly, the optimal solution in regime 1 is essentially identical to that of (\ref{eq:shapefit/kick}). However, it is generally not the optimal solution if we permit regime 2, which solves the original (\ref{eq:LUD}) without the $\gamma\bm{t}^S$ restriction. As a consequence, the total residual may be further reduced by adjusting the scale of each residual term individually without respecting the overall global shape. In particular, those $i{-}j$ terms representing large baselines often have larger residuals as any slight deviation of the solution from the observation is scaled by the baseline magnitude. As the corresponding $d_{ij}$'s are less likely to have  reached the lower bound $c$ (since the optimal $d_{ij}{=}\min(\left\langle\hat{\bm{t}}_j-\hat{\bm{t}}_i,\bm{v}_{ij}\right\rangle,c)$ for the solution $\hat{\bm{t}}$), it thus pays to scale those larger baselines $||\bm{t}_j{-}\bm{t}_i||_2$ and $d_{ij}$'s down by shrinking camera $i$ and $j$ closer together as long as the reduction in residuals at these $i{-}j$ terms can more than make up for the increase in residuals in other terms. The best solution would therefore exhibit a partial shrinking effect, with the overall shape of the camera configuration squashed.
\begin{figure}
	\setlength{\abovecaptionskip}{-0.3cm}
	\setlength{\belowcaptionskip}{-0.55cm}
	\vspace{-0.1cm}
	\begin{center}
		\includegraphics[width=0.7\linewidth, trim = 0mm 5mm 00mm 0mm, clip]{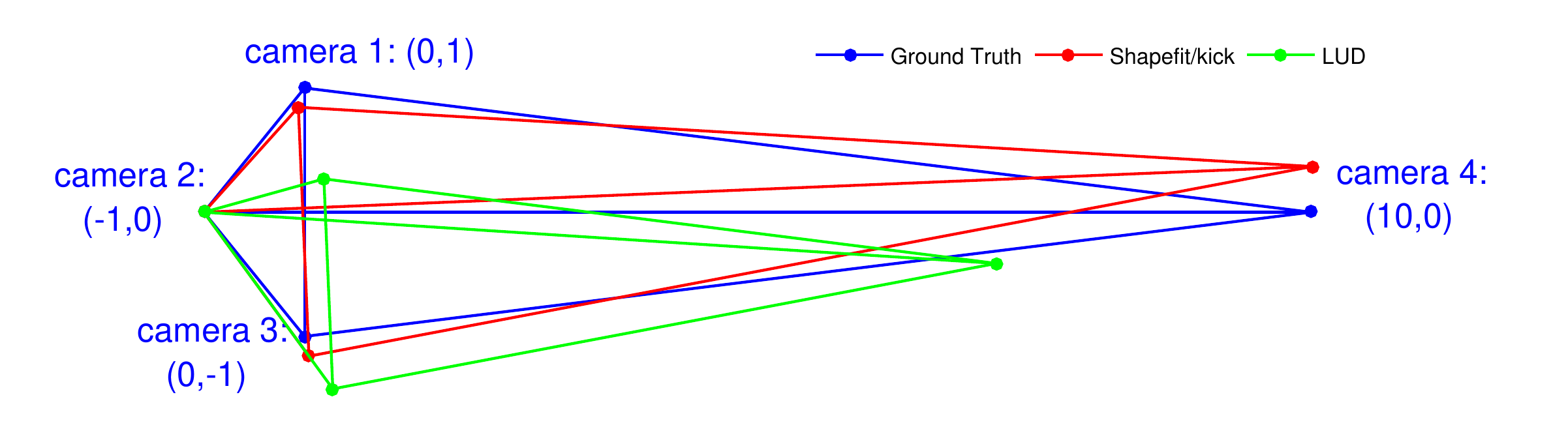} \\
		\includegraphics[width=1\linewidth, trim = 0mm 228mm 150mm 68mm, clip]{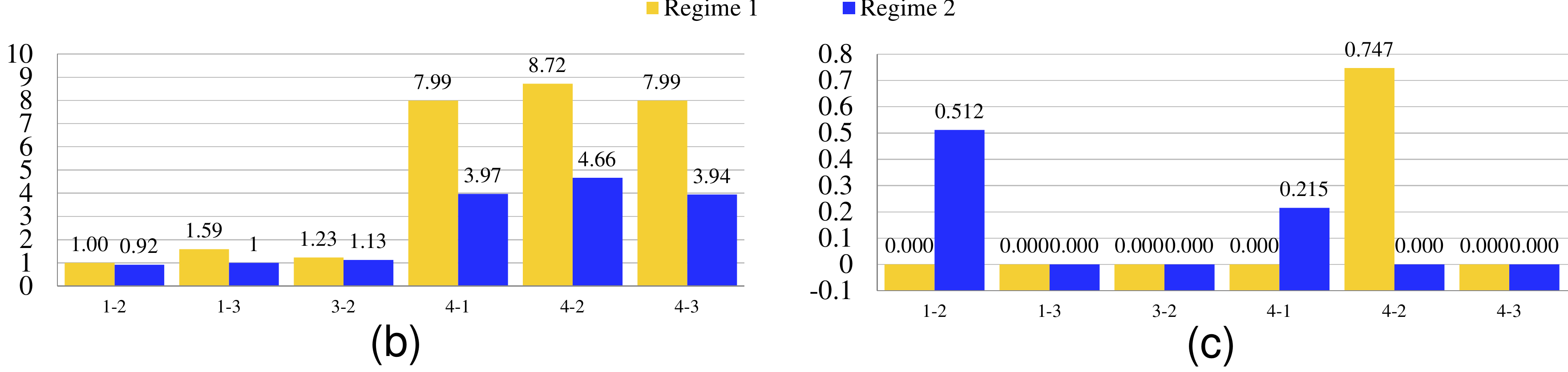}
	\end{center}
	\caption{A toy example illustrating the squashing effect. (a) True and estimated camera locations (for better shape comparison, the three sets of camera configurations are aligned at camera 2 and normalized to have the same perimeter in the triangle formed by camera 1-3). 
		(b)-(c) $\lVert\bm{t}_j{-}\bm{t}_i\rVert_2$ and $\lVert\bm{t}_j{-}\bm{t}_i{-}d_{ij}\bm{v}_{ij}\rVert_2$ plotted against $i{-}j$, which denotes the edge between camera $i$ and $j$.}
	\label{fig:toyexampleLUD}
\end{figure}
%and the optimization in (\ref{eq:LUD}) is conceptually separated into the following two phases. First, it tries to optimize the objective by overall scaling up or down the $\bm{t}^S$, i.e. the solution is restricted to be of the form $\gamma\bm{t}^S$ where $\gamma$ is a scale to be optimized. Clearly, this phase remains the solution unchanged essentially. However, after this phase, the total residual may still be reduced by adjusting each residual term individually. Especially, it pays to scale those larger baseline $||\bm{t}_j-\bm{t}_i||_2$ and $d_{ij}$ down by pushing camera $i$ and $j$ closer to each other as long as the reduction in residual at these locations can more than make up for the increase in residuals in other locations. This causes a partial shrinking effect, especially in the presence of largely varying baselines, and distorts the geometric configuration of the camera positions. 
We give a concrete toy example to illustrate how disparate baselines would exacerbate this tendency. For simplicity, let us look at a four-camera 2D case where the true camera locations are at (0, 1), (-1, 0), (0, -1) and (10, 0), and all their pairwise relative directions are observed with a $3^\circ$ noise. We first visualize the solutions obtained from (\ref{eq:LUD}) and from (\ref{eq:shapefit/kick}) together with the ground truth in Fig.~\ref{fig:toyexampleLUD}(a). As can be seen, the distant camera 4 gravitates significantly towards cameras 1-3 in the solution of (\ref{eq:LUD}). We also plot the pairwise camera distances and the residuals of the best solutions (with $c{=}1$) from regime 1 (optimized by linearly searching $\gamma$) and 2 in Fig.~\ref{fig:toyexampleLUD}(b)\&(c), respectively. Fig.~\ref{fig:toyexampleLUD}(b) corroborates what we said above: compared to the solution of regime 1, regime 2 tends to pull those well-separated camera pairs closer together. Note that for those camera pairs that have short baselines, their corresponding $d_{ij}$ has already reached the lower bound, and thus the $\bm{t}_i$ and $\bm{t}_j$ cannot shrink in tandem with those of the well-separated cameras without $\lVert\bm{t}_j{-}\bm{t}_i\rVert_2$ deviating too far from the $d_{ij}$,  leading to squashing effect in the camera configuration. Referring to Fig.~\ref{fig:toyexampleLUD}(c), first note that in this toy example the solutions of both regime 1 and 2 agree with most of the observations exactly; we attribute this to the fact that the objectives of (\ref{eq:LUD}) and (\ref{eq:shapefit/kick}) are actually based on a group-sparsity term \cite{yuan2006model,nie2010efficient}, i.e. $L_{2,1}$ norm, which favors sparse residuals. Note also that regime 2 achieves a lower total residual (0.727) compared to regime 1 (0.747) by suppressing the large residual (the lone yellow peak) at the expense of the small residuals, validating our analysis above. 

This observation is generally useful and applicable to other methods where such lower bound exists (e.g. \cite{moulon2013global}), or latter works based on LUD framework (e.g. \cite{Sengupta_2017_CVPR,theia-manual}). 
%We also note that the RevisedLUD and BATA differ from each other only slightly by the position of $d_{ij}$, which nevertheless leads to two rather different geometric quantities being minimized.

%% file: Section/sec_experiment.tex
\vspace{-0.1cm}
\section{Experiments}
\label{sec:expe}
%\vspace{-0.1cm}
%In this section, we demonstrate experimental results on both synthetic and the real 1DSfM benchmark dataset \cite{wilson2014robust}.  %the superiority of our framework compared to state-of-the-art methods via experiments 
%\vspace{-0.5cm}
\subsection{Synthetic Data Experiments}
\vspace{-0.1cm}

We first study the performance of different methods on synthetic data. To synthesize the view graph, we first generate the ground-truth camera locations $\bar{\bm{t}}_i, \forall i \in V$, by drawing i.i.d. samples from  $N(0,I_{3\times 3})$. Denoting the number of cameras as $n$ (set as 200 here), the pairwise edges $E$ are then drawn randomly from the Erd\H os-R\' enyi model $\mathcal{G}(n,p)$, meaning each edge is observed with probability p, independently of all other edges. We then perturb the observed pairwise directions to mimic the effect of noises and outliers. As opposed to \cite{ozyesil2015robust,goldstein2016shapefit} where Gaussian noises are added to the endpoint of the direction vector followed by a normalization to be of unit norm, we directly add noise to the orientation of the pairwise direction; we believe this to be more reflective of the actual perturbation.
Specifically, we obtain each corrupted pairwise direction $\bm{v}_{ij}$ as follows,

\vspace{-0.6cm}
\begin{equation}
\hspace{-0.2cm}
\bm{v}_{ij} = \left\{
\begin{aligned}
& \bm{v}_{ij}^u, &  \text{with probability q}, \\
& R(\sigma\theta_{ij}^g,\bm{h}_{ij}^u)\frac{\bar{\bm{t}}_j-\bar{\bm{t}}_i}{||\bar{\bm{t}}_j-\bar{\bm{t}}_i||}, & \text{otherwise};
\end{aligned}
\right.
\end{equation}
\vspace{-0.4cm}

\noindent where $\bm{v}_{ij}^u$ and $\bm{h}_{ij}^u$ are i.i.d. unit random vectors drawn from uniform distribution on the unit sphere and the orthogonal complement of the span of $\frac{\bar{\bm{t}}_j-\bar{\bm{t}}_i}{\lVert\bar{\bm{t}}_j-\bar{\bm{t}}_i\rVert_2}$, respectively. $\theta_{ij}^g$ is drawn from i.i.d. $N(0,1)$ and $\sigma$ is a scale controlling the noise level. $R(\sigma\theta_{ij}^g,\bm{h}_{ij}^u)$ is a rotation matrix around the aixs $\bm{h}_{ij}^u$ for an angle $\sigma\theta_{ij}^g$ (counter-clockwise). Like \cite{ozyesil2015robust}, we use the normalized root mean square error (NRMSE) to evaluate the accuracy: $NRMSE{=}\sqrt{\sum_{i\in V}\lVert\hat{\bm{t}}_i-\bar{\bm{t}}_i\rVert_2^2}$,
%\vspace{-0.2cm}
%\begin{equation}
%NRMSE = \sqrt{\sum_{i\in V}\lVert\hat{\bm{t}}_i-{\bm{t}}_i\rVert^2},
%\end{equation}
%\vspace{-0.4cm}
where $\hat{\bm{t}}_i, \forall i \in V$ is the set of estimated locations. Both $\hat{\bm{t}}$ and $\bar{\bm{t}}$ are centralized and normalized, i.e. $\sum_{i\in V}\hat{\bm{t}}_i = \bm{0}$, $\sum_{i\in V}\lVert\hat{\bm{t}}_i\rVert_2^2 = 1$ and the same is true of $\bar{\bm{t}}$.

\begin{figure}
 \setlength{\abovecaptionskip}{-0.4cm}
 \setlength{\belowcaptionskip}{-0.4cm}
 \vspace{-0.1cm}
\begin{center}
   \includegraphics[width=0.85\linewidth, trim = 10mm 32mm 7mm 8mm, clip]{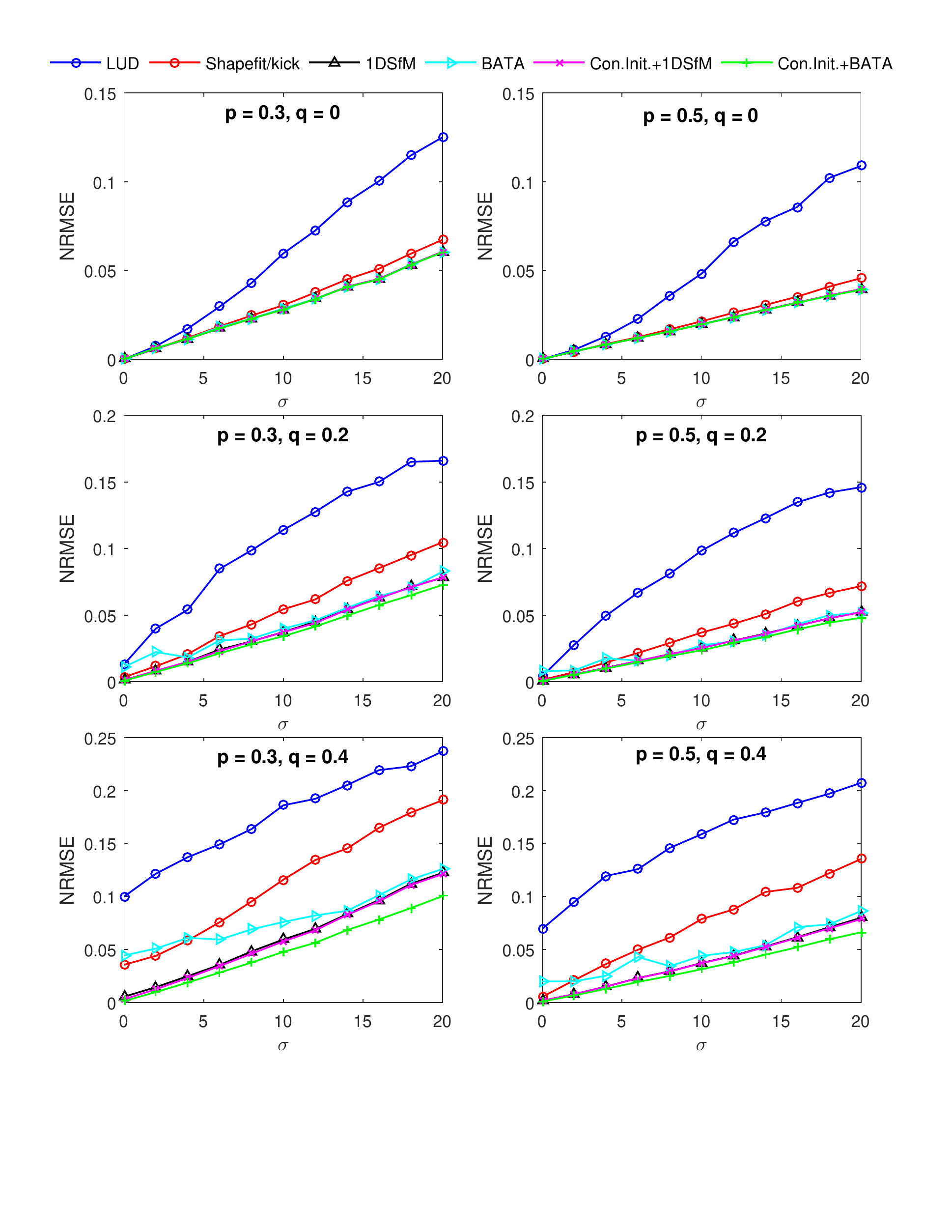}
\end{center}
   \caption{NRMSE from different methods under different view graph setup $(p,q)$ and noise level $\sigma$.}
\label{fig:rfe}
\end{figure}

\begin{figure}
	\setlength{\abovecaptionskip}{-0.4cm}
	\setlength{\belowcaptionskip}{-0.6cm}
	\begin{center}
		\includegraphics[width=0.83\linewidth, trim = 10mm 0mm 8mm 0mm, clip]{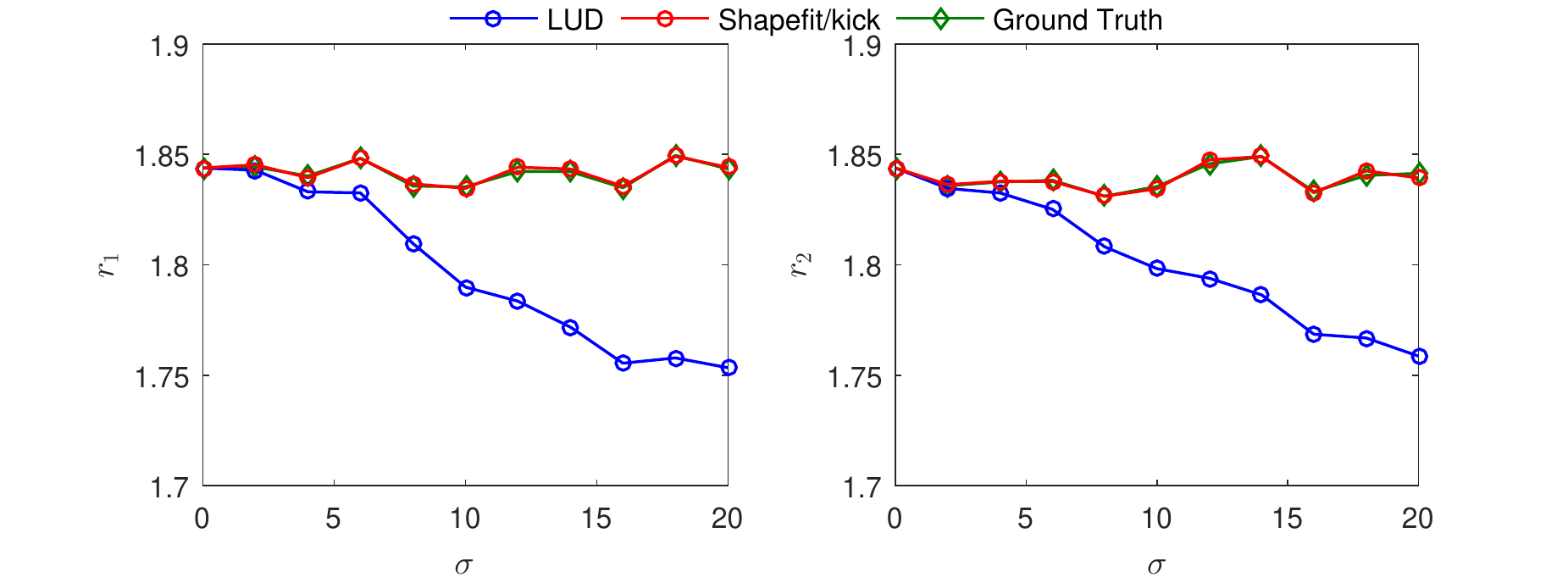}
	\end{center}
	\caption{$r_1$ and $r_2$ from the $(p,q){=}(0.3,0.2)$ case.}
	\label{fig:partialshrink}
\end{figure}

We compare the performance of different objectives including LUD \cite{ozyesil2015robust}, Shapefit/kick \cite{goldstein2016shapefit},  the nonlinear objective
from 1DSfM \cite{wilson2014robust}, and BATA. We follow 1DSfM to use Huber as the robust scheme for BATA.  For 1DSfM and BATA, we evaluate both random initialization and initialization from RevisedLUD. In the latter, we run a few iterations (IRLSiter=10 \& BCDiter=1) of na\"{\i}ve IRLS for RevisedLUD to bootstrap the 1DSfM and BATA (the results are denoted as ``Con.Init.+1DSfM" and ``Con.Init.+BATA"). We use Ceres \cite{ceres-solver} for the optimization of 1DSfM. LUD and Shapefit/kick are optimized by IRLS and ADMM. All methods are run until they are well converged (we fix the number of iterations for BATA as IRLSiter=20 \& BCDiter=5). 

We investigate the performance under six combinations of observation ratio $p$ and outlier ratio $q$, each with increasing noise level $\sigma$, as shown in Fig.~\ref{fig:rfe}. The results are averaged over 20 independently generated view graphs. As we can see, angle-based objective functions generally achieve lower NRMSE compared to the magnitude-based counterparts. In particular, both 1DSfM and BATA can achieve better performance than LUD and Shapefit/kick even with random initialization, with the difference more notable under larger noises. Additionally, we observe that a good initialization is not important for 1DSfM here.
% and BATA in those more well-conditioned configurations with higher $p$ and lower $q$, and vice versa. 
We also observe that although 1DSfM and BATA tend to perform equally well in the outlier-free configurations, BATA, if bootstrapped with a good initialization, achieves higher accuracies when the camera configuration becomes increasingly ill-conditioned with lower $p$ and higher $q$, e.g. the bottom-left case. We attribute this to the leveling off in the objective of BATA. Next, we demonstrate the partial shrinking bias caused by the lower-bound constraint in LUD. Under increasing noise, we monitor the following two ratios
\vspace{-0.2cm}
$$r_1{ =} pct(\{S_{ij}|ij{\in} E\},75)/pct(\{S_{ij}|ij{\in} E\},25),$$
\vspace{-0.5cm}
 $$r_2{=} pct(\{\lVert\bm{t}_i\rVert_2|i\in V\},75)/pct(\{\lVert\bm{t}_i\rVert_2|i\in V\},25),$$ 
 to measure the extent of the partial shrinking bias, where $S_{ij}{=}\lVert\bm{t}_j{-}\bm{t}_i\rVert_2$ and $pct(\bm{a},b)$ denotes the $b$-th percentile of $\bm{a}$. We plot the result for the $(p,q){=}(0.3,0)$ case in Fig.~\ref{fig:partialshrink} and leave other cases to the \textit{supp. material}. As can be seen, compared to those in Shapefit/kick, both $r_1$ and $r_2$ in LUD become increasingly smaller than the ground truth under increasing noise level, indicating a squashing effect.

\begin{figure}
	\setlength{\abovecaptionskip}{-0.3cm}
	\setlength{\belowcaptionskip}{-0.45cm}
	\centering
	\vspace{-0.2cm}
	\begin{center}	
		\centering
		\includegraphics[width=0.32\linewidth, trim = 35mm 0mm 50mm 10mm, clip]{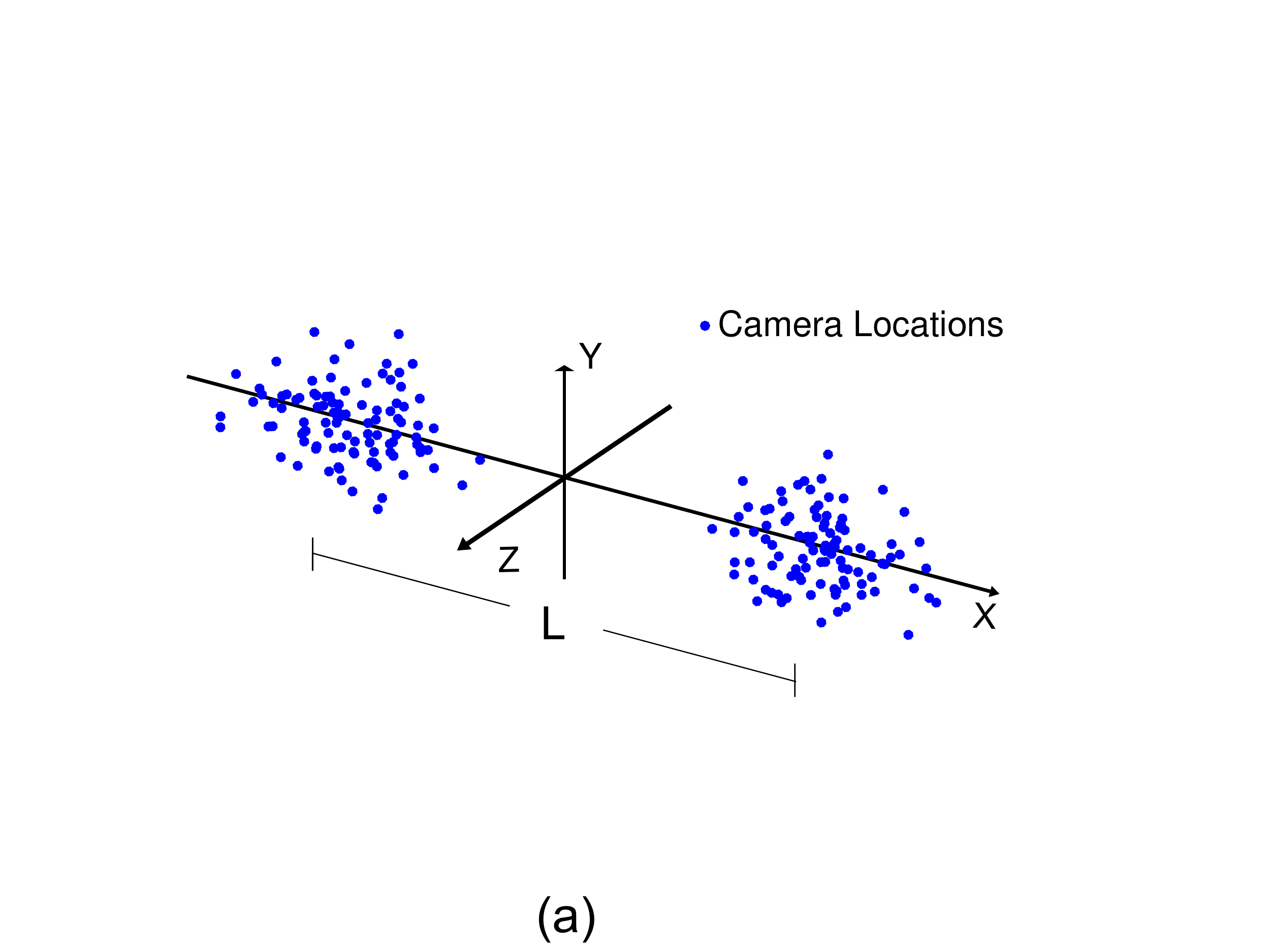}
		\includegraphics[width=0.66\linewidth, trim = 16mm 0mm 25mm 85mm, clip]{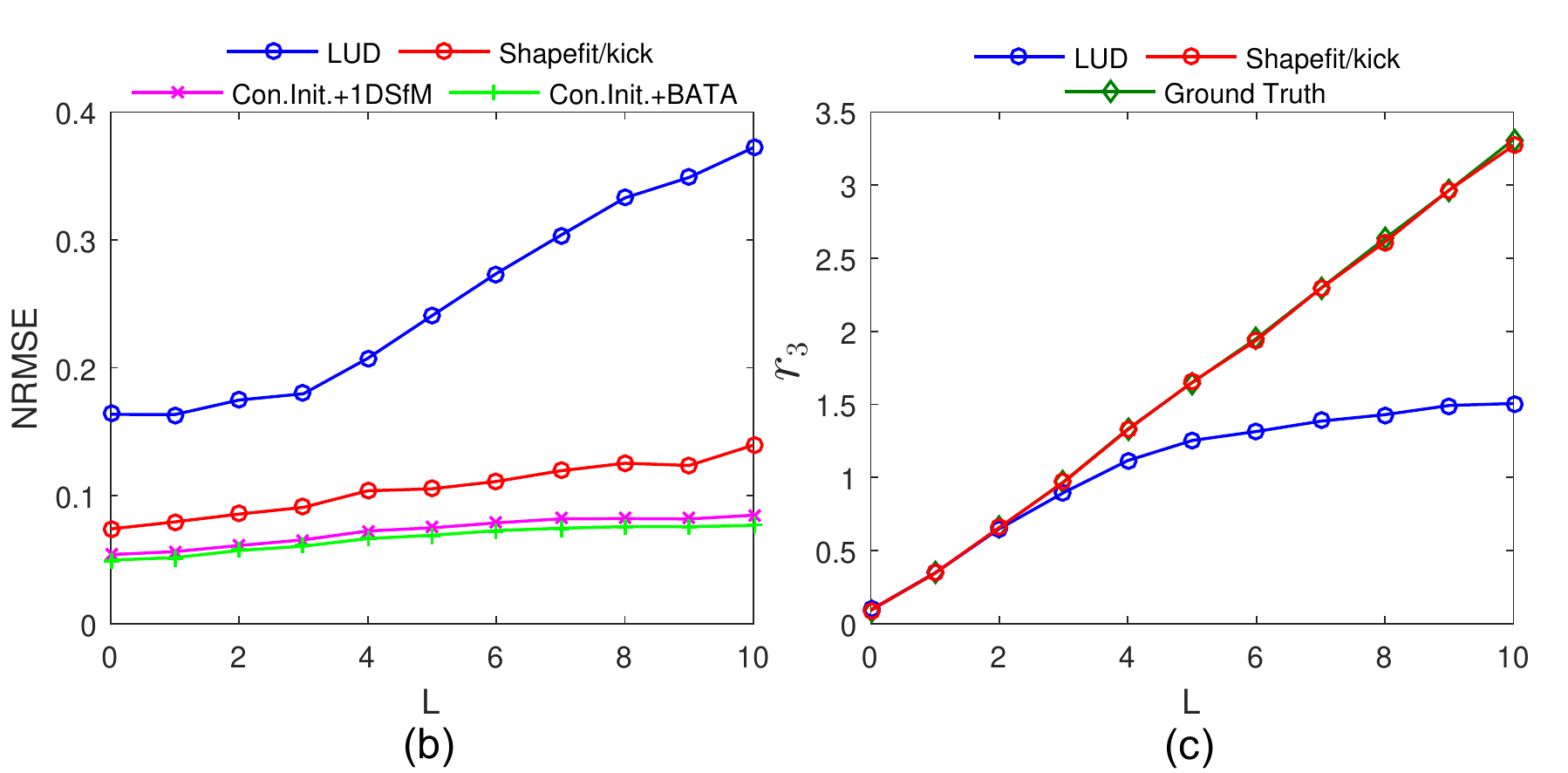}
	\end{center}
	\caption{(a) Illustration of the two camera clusters. (b)-(c) NRMSE and $r_3$. Both are averaged over 20 trials.}
	\label{fig:twosphere}
\end{figure}

Finally, we test the sensitivity of the algorithms to an unbalanced distribution in the baseline magnitudes. For this purpose, the true camera locations are sampled from two separate clusters $N([-L/2,0,0],I_{3\times3})$ and $N([L/2,0,0],I_{3\times3})$, as illustrated by the dots in Fig.~\ref{fig:twosphere}(a). The larger the $L$ is, the more significant the unbalance in the baseline magnitudes is. We fix $(p,q){=}(0.3,0.2)$ and $\sigma{=}10$, and increase the value of $L$ from 0 to 10. For each solution, we compute NRMSE\footnote{Here, we centralize and normalize two clusters separately to avoid the inherent decreasing of NRMSE while increasing $L$.} and the ratio $r_3=l_{12}/(l_1+l_2)$.  $l_{12}$ denotes the distance of the two cluster centers. $l_1$ and $l_2$ denote the median value of the set of distances from each point to their centers in the two clusters respectively. Note that $r_3$ explicitly measures the squashing effect caused by the different shrinking rates experienced by the longer inter-cluster baselines versus that of the shorter intra-cluster baselines. As shown in Fig.~\ref{fig:twosphere}(b)-(c), under increasing $L$, the NRMSE's from the two magnitude-based methods, especially LUD,
% Shapefit/kick, and especially, LUD, 
increase more significantly, meaning that they are more susceptible to the disparate baselines; it is also clear that $r_3$ from LUD decreases substantially compared to the ground truth, indicating a significant squashing effect.
% This corroborates our analysis in Sec.~\ref{sec:objfunc} and Sec.~\ref{sec:whatLUD}. 

\subsection{Real Data Experiments}
\begin{figure*}[htpb]	
	\setlength{\abovecaptionskip}{-0.4cm}
	\setlength{\belowcaptionskip}{-0.35cm}
	\vspace{-0.4cm}
	\begin{center}
		\includegraphics[width=0.99\linewidth, trim = 0mm 100mm 5mm 22mm, clip]{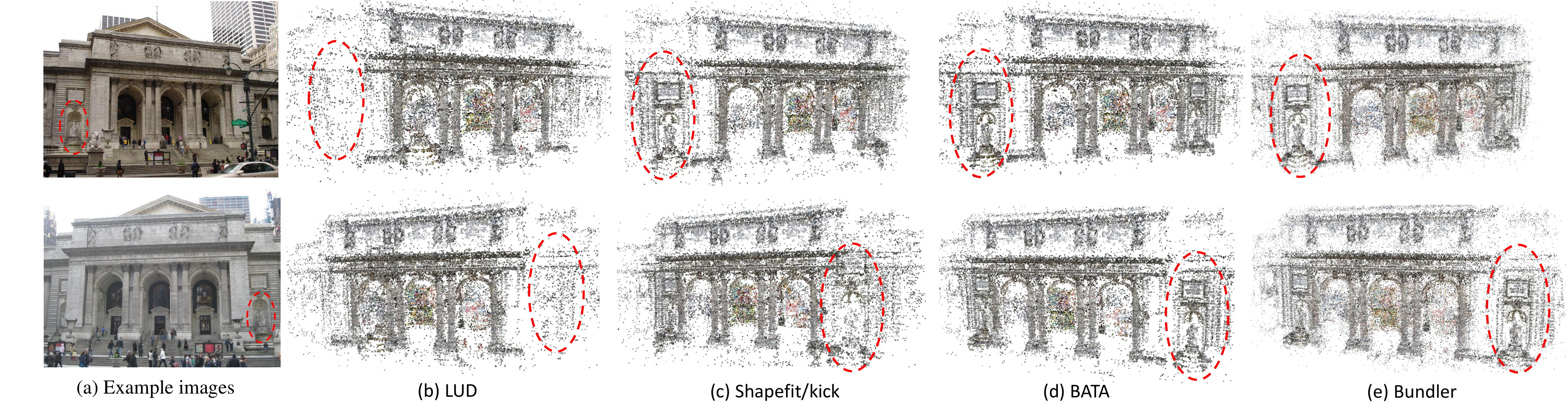}
	\end{center}
	\caption{Visualization of the point clouds after final BA on the NYC Library data. (a) depicts two images of the scene, with the red ellipses highlighting two sculptures. (b)-(e) show the resultant point clouds by the respective methods.}
	\label{fig:pointcloud}
\end{figure*}

\begin{table*}\footnotesize	
	\setlength{\tabcolsep}{2.1pt}
	\setlength{\abovecaptionskip}{-0.25cm}
	\setlength{\belowcaptionskip}{-0.5cm}
	\begin{center}
		\scalebox{0.95}{
			\hspace{-0.15cm}
			\begin{tabular}{|c|c||c|c||c|c||c|c||c|c||c|c|c|c|c|c||c|c|c|c|c|c|c|c|}
				\hline		
				\multicolumn{2}{|c||}{\multirow{2}{*}{Data}} &			
				\multicolumn{2}{c||}{\multirow{2}{*}{1DSfM\cite{wilson2014robust}}} &
				\multicolumn{2}{c||}{\multirow{2}{*}{LUD\cite{ozyesil2015robust}}} &
				\multicolumn{2}{c||}{\multirow{2}{*}{Shapefit/kick\cite{goldstein2016shapefit}}} &
				\multicolumn{2}{c||}{\multirow{2}{*}{Cui\cite{cui2015global}}} &
				\multicolumn{14}{c|}{BATA}\\			
				\cline{11-24} 
				\multicolumn{2}{|c||}{}& \multicolumn{2}{c||}{} &\multicolumn{2}{c||}{}& \multicolumn{2}{c||}{}& \multicolumn{2}{c||}{} & \multicolumn{3}{c|}{R.I. w/o R.}  & \multicolumn{3}{c||}{R.I. w R.}  & \multicolumn{2}{c|}{Con. Init.} & \multicolumn{3}{c|}{w/o Rot.} & \multicolumn{3}{c|}{w Rot.} \\
				\hline
				Name &  $N_c$ &$\tilde{e}$  & $\bar{e}$  &$\tilde{e}$  & $\bar{e}$   & $\tilde{e}$  & $\bar{e}$  & $\tilde{e}$  & $\bar{e}$  & $\tilde{e}$  & $\bar{e}$ & \#iter  & $\tilde{e}$  & $\bar{e}$  & \#iter
				& $\tilde{e}$  & $\bar{e}$   & $\tilde{e}$  & $\bar{e}$ & \#iter & $\tilde{e}$  & $\bar{e}$ & \#iter \\
				\hline
				{Piccadilly (PIC)}      & 2508 &  \hspace{0.05cm}1.4 \hspace{0.05cm} & 6e3  &  \hspace{0.1cm}-\hspace{0.1cm}        &-      &  \hspace{0.2cm}1.2 \hspace{0.2cm}  & 15   & 1.3   &  2.5  & 4.3  & 13.8 &100 & 1.5  & 7.5    &78 &3.0   & 5.0   &1.0  &5.2  &100 &\bf{1.0}  &4.2 & 57\\ 
				\hline 
				{Union Sq. (USQ)}      & 930  &  5.0  & 2e3 & -        &-      & 8.9   & 47   & 5.5   &  12.7 & 7.2  & 15.4 &100 & 6.3  & 14.9   &100  &6.2    & 11.9   &4.3   &12.4 &100 &\bf{4.3}  &12.3 &100\\
				\hline
				{Roman For. (ROF)}       & 1134 & 3.2 & 2e4  & -        &-      & 4.3   & 25   & 2.9   &  9.4  & 3.2  & 24.2 &100 & 2.4  & 23.7   &100  &9.4   & 20.8   &1.6  &9.9 &100 &\bf{1.6}  &16.3 &88\\
				\hline
				{Vienna Cath. (VNC)}  & 918  & 2.1  & 3e4 & \hspace{0.01cm} 5.4 \hspace{0.01cm}     &10     & \bf{1.9}   & 11   &2.7    &  5.9  & 2.5  & 18.2 &100 & 2.0  & 16.5  &77  &6.1    &13.1    &1.9   &12.1  &99 &\bf{1.9}  &13.3 &74\\
				\hline
				{Piazza Pop. (PDP)} & 354  & 3.0  & 54.2 & \bf{1.5}      &5      & 3.6   & 5.9   &{2.0}    & 2.7   & 2.1  & 11.0 &100 & 1.9  & 8.9     &96 &1.4    &6.5    &1.7  &6.7 &100 &4.2  &6.2 &62\\
				\hline
				{NYC Library (NYC)}      & 376  & 0.9  & 1e4 & 2.0      &6      & 1.4   & 162   &0.8    & 1.9  &1.0  & 9.1 &100 &0.7  &6.6  &87 &1.1    &3.3    &0.7 &3.2  &83 &\bf{0.6}  &2.7 &61\\
				\hline
				{Alamo (ALM) }           & 627  & 0.8  & 1e3 & \bf{0.4} & 2     & 0.9   & 5.0   & 0.5   & 2.0  & 0.6 & 7.2 &85 & 0.6 & 6.2     &49  &1.8    &3.9   &0.5 &3.4 &61  &{0.6}  &3.3 &40\\
				\hline
				{Metropolis (MDR)}        & 394  & 3.8  & 6e4 & \bf{1.6} & 4     & 6.0   & 81   &2.7    & 10.6  & 3.6 & 34.1 &100 & 2.1 & 24.5    &85   &4.5    &15.7   &4.0 &15.3 &97 &{1.8} &12.1 &64\\
				\hline
				{Yorkminster (YKM) }     & 458 & 1.7   & 1e4 & 2.7         & 5     & -    & -    &2.3   &5.7 & 1.2  & 15.9 &100 & 1.0  & 15.2   &98 &4.4    &12.8   &1.3 &8.4  &100 &\bf{0.9}  &8.0 &85\\
				\hline
				{Montreal N.D. (MND)}     &474  & 0.8   & 5e4 &0.5        & 1     & 0.8  & 1.7  &{0.4} &0.7 & 1.0  & 4.0  &94 &  0.5 & 1.8   &69     &1.0    & 1.7   &0.4 &0.8 &64  &\bf{0.3}  &0.7 &45\\
				\hline
				{Tow. London (TOL)}   &508  & 3.1   & 6e3  & 4.7       & 20    & 2.3    & 164    &\bf{1.9} &11.2  & 2.5  & 25.2 &100 & 2.3  & 18.5 &99  &5.1    &22.9    &2.1 &13.5 &100  &{2.2}  &16.0 &100\\
				\hline
				{Ellis Island (ELS)}      & 247 & 1.8  & 9.8  & -         & -     & 1.9    & 12    &2.5 &5.5  &1.6  & 22.5 &97 &1.5 & 15.8   &53  &2.2    &9.7    &1.4 &11.1 &80  &\bf{1.5}  &13.4 &39\\
				\hline
				{Notre Dame (NOD)}        & 553 & \bf{0.2} & 1e3  & 0.3       &0.8    & \bf{0.2}    & 1.5  &\bf{0.2} & 0.6 & 0.4   &5.7  &100 & 0.2   &4.5  &76 &3.1    &4.1    &0.3 &1.8 &96 &\bf{0.2}  &2.1 &70\\
				\hline		
      			{Trafalgar (TFG)}        & 5433 & 5.0 & 3e3  & -      &-    & -    & -  &5.4 & 8.9 & 6.2   &23.2  &100 & 4.1   &18.8  &92 &8.8    &14.7    &3.9 &12.2 &89 &\bf{3.4}  &11.7 &65\\
				\hline		
			\end{tabular}
		}
	\end{center}
	\caption{Comparison of the accuracy of different methods in real data. $N_c$ is the number of cameras in the view graph. $\tilde{e}$ and $\bar{e}$  respectively denote the median and mean distance error in meter unit. \#iter denotes the number of outer iterations (the value of $n$ in Algo.~\ref{algo:irlsbcd}) required for convergence, bounded by 100. `-' indicates that the result is not available from the corresponding paper.} 	
	\label{tab:accuracycomparison}
\end{table*}

We now present the results on real unordered photo collections provided by the 1DSfM dataset \cite{wilson2014robust} (see Fig.~\ref{fig:pointcloud}(a) for examples). The raw largest connected view graph released along with the dataset is used as our input. Similar to \cite{wilson2014robust,goldstein2016shapefit,cui2015global,ozyesil2015robust}, we apply the method of \cite{chatterjee2013efficient} to perform rotation averaging. To quantitatively evaluate the quality of a translation averaging estimate, it is compared with the gold standard output by Bundler \cite{snavely2006photo}; the two sets of camera positions are robustly registered using the codes of \cite{wilson2014robust}.

We evaluate the performance of a few different setups under BATA to understand its behavior. The first case of interest is to simply run BATA from random initialization in two settings, without or with rotation involved in the IRLS re-weighting (denoted as ``R.I.~w/o~R." and ``R.I.~w~R."). Next, we use as initialization the moderately accurate output of a convex algorithm: to this end,  we run a few rotation-assisted IRLS iterations (IRLSiter=50 \& BCDiter=1) of RevisedLUD (denoted as ``Con.~Init."). Again, BATA is run in the above two settings (denoted as ``w/o Rot." and ``w Rot."). We set IRLSIter=100 \& BCDIter=5 with convergence condition being $\lvert f^c{-} f^l\rvert/f^l{<}10^{-5}$, where $f^l$ and $f^c$ are the objective values of two consecutive iterations. All results are averaged over 20 trials.

 \begin{table*}\footnotesize	.
 	\setlength{\abovecaptionskip}{+0.1cm}
 	\setlength{\belowcaptionskip}{-0.2cm}
 	\setlength{\tabcolsep}{3pt}		
 	\hspace{-0.1cm}	 	 
 	\begin{minipage}[b]{0.3\linewidth}  			 	
 		%\centering
 			\scalebox{1}{
 		\hspace{+0.0cm}
 		\begin{tabular}[H]{|@{ }c@{ }||c|c||c|c|}
 			\hline
 			\multicolumn{1}{|c||}{\multirow{1}{*}{Data}} & \multicolumn{2}{c||}{\multirow{1}{*}{LUD}} & \multicolumn{2}{c|}{\multirow{1}{*}{Shapefit/kick}}\\ 
 			%& \multicolumn{3}{c|}{} & \multicolumn{2}{c|}{} \\
 			%& \multicolumn{3}{c|}{} & \multicolumn{2}{c|}{} \\
 			\hline
 			Name  &  $r_1$ & $r_2$ & $r_1$ & $r_2$\\ 
 			\hline
 			PIC  &  2.26 & 2.43 & 2.71 & 2.81\\
 			\hline
 			USQ &  2.27& 2.60 & 6.54 & 3.08\\
 			\hline
 			ROF & 2.54& 2.19 & 4.62 & 2.57\\
 			\hline
 			VNC & 2.42 & 2.85 & 2.73 & 2.63\\
 			\hline
 			PDP & 2.48& 2.57 & 2.87 & 2.33 \\
 			\hline
 			NYC & 2.33& 2.22 & 2.69 & 2.34\\
 			\hline
 			ALM  & 2.24 & 2.65 & 2.55 & 2.82\\
 			\hline
 			MDR  & 2.28 & 2.22  & 6.95 & 10.2\\
 			\hline
 			YKM  & 2.46& 2.24 & 3.21 & 2.86\\
 			\hline
 			MND & 2.83 & 2.13 & 3.65 & 1.74\\
 			\hline
 			TOL & 2.41 & 2.73 & 3.12 & 2.38\\
 			\hline
 			ELS & 1.86 & 2.26 & 2.09 & 3.19\\
 			\hline
 			NOD & 2.43 & 2.34 & 2.96 & 2.58\\
 			\hline
 			TFG & 2.27 & 2.24 & 2.63 & 3.03\\
 			\hline
 		\end{tabular}  }
 	 	\caption{Comparison of $r_1$ and $r_2$ \\from LUD and Shapefit/kick.} 
 		\label{tab:realdataratio}	
 	\end{minipage}%	 
 	\hspace{-0.6cm}
 	\begin{minipage}[b]{0.7\linewidth}		
 		\centering 	 	
 		\scalebox{1}{	
 		\begin{tabular}[H]{|@{ }c@{ }||c|c|c||c|c|c||c|c|c||c|c|c||c|c|c|}
 			\hline
 			\multicolumn{1}{|c||}{\multirow{1}{*}{Data}} & \multicolumn{3}{c||}{\multirow{1}{*}{1DSfM\cite{wilson2014robust}}} & \multicolumn{3}{c||}{\multirow{1}{*}{LUD\cite{ozyesil2015robust}}} & \multicolumn{3}{c||}{\multirow{1}{*}{Shapefit/kick\cite{goldstein2016shapefit}}} & \multicolumn{3}{c||}{\multirow{1}{*}{Cui\cite{cui2015global}}} & \multicolumn{3}{c|}{\multirow{1}{*}{BATA}}\\
 			%& \multicolumn{3}{c|}{} & \multicolumn{2}{c|}{} \\
 			%& \multicolumn{3}{c|}{} & \multicolumn{2}{c|}{} \\
 			\hline
 			Name  &  $T_p$ & $T_t$ & $T_\Sigma$&  $T_p$ & $T_t$ & $T_\Sigma$&  $T_p$  & $T_t$ & $T_\Sigma$ &  $T_p$  & $T_t$ & $T_\Sigma$& $T_{ini}$  & $T_t$ & $T_\Sigma$\\
 			\hline
 			PIC  &  122 & 366 &488 & - & - & -& 424 & 40 & 464 & 207 &121 & 328  & 52.9  &60.6 & 113.5 \\
 			\hline
 			USQ &  20& 75 &95 & - & - & - & 24 & 3.7 & 27.7 & 35 &6 & 41 &  2.5 & 7.5 & 10.0\\
 			\hline
 			ROF & 40& 135 &175 & - & - & - & 52 & 9.5 & 61.5 & 99 &32 & 131 &8.1 &20.9 & 29.0\\
 			\hline
 			VNC & 60 & 144 &204 & 265 & 255 & 520 & 66 & 8.2 & 74.2  &102 &15 & 117 & 12.6 & 17.2 & 29.8 \\
 			\hline
 			PDP & 9& 35 & 44  & 18 & 35 & 53 & 4.6 & 1.9 & 6.5 &40 &3 & 43 &2.0  & 2.2 & 4.2\\
 			\hline
 			NYC & 13& 54 & 67 & 18 & 57 & 75 & 8.6 & 2.2 & 10.8 & 34 &4 & 38 & 1.7 &2.1 & 3.8\\
 			\hline
 			ALM  & 29 & 73 & 102 & 96 & 186 & 282 & 16 & 11 & 27 & 67 &11 & 78 & 11.1 & 13.0 & 24.1\\
 			\hline
 			MDR  & 8 & 20  & 28 & 13 & 27 & 40 & 6.9 & 2.4 & 9.3  &27 &4 & 31  & 2.0 & 2.4 & 4.4 \\
 			\hline
 			YKM  & 18& 93 & 111 & 33 & 51 & 84 & - & - & -  & 41 &5 & 46 & 2.4 & 6.3 & 8.7\\
 			\hline
 			MND & 22 & 75 & 97 & 91 & 112 & 203 & 15 & 3.5 & 18.5 &57 &5 &62 &5.4  & 4.1 & 9.5 \\
 			\hline
 			TOL & 14 & 55 & 69 & 23 & 41 & 64 & 15 & 2.8 & 17.8 & 46 &6 & 52& 2.1 & 4.4 & 6.5\\
 			\hline
 			ELS & 7 & 13 & 20 & - & - &-  & 2.9 & 1.4 & 4.3 & 34 &3 &37 & 1.4 & 1.0 & 2.4\\
 			\hline
 			NOD & 42 & 59 & 101 & 325 & 247 &572  & 23 & 7.1 &  30.1 & 61 &9 & 70 &11.7  &11.7 & 23.4\\
 			\hline
 			TFG & - & - & - & - & - &-  & - & - &  - & 441 & 583 & 1024 & 168.9  &389.1 & 558.0\\
 			\hline
 		\end{tabular}  }
 	 	\caption{Comparison of running time in seconds. $T_p$, $T_{ini}$, $T_t$, and $T_\Sigma$ respectively denote the preprocessing time, initialization time, translation averaging time, and total time.} 	
 	 	\label{tab:timing}	
 	\end{minipage} 
 	\vspace{-0.35cm}
 \end{table*}

\begin{figure}
	\setlength{\abovecaptionskip}{-0.3cm}
	\setlength{\belowcaptionskip}{-0.8cm}
	\vspace{-0.2cm}
	\begin{center}
		\includegraphics[width=0.75\linewidth, trim = 6mm 0mm 10mm 0mm, clip]{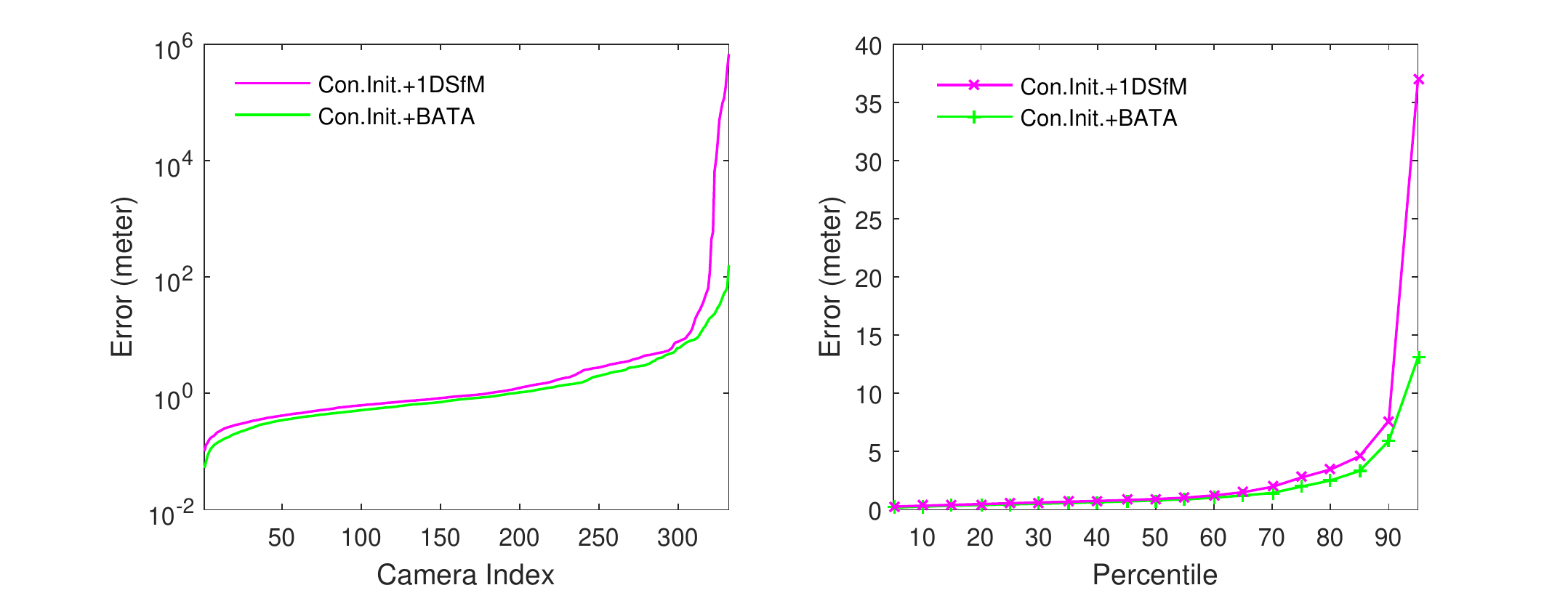}
	\end{center}
	\caption{(a) Errors distribution (sorted in increasing order) for all cameras. (b) Errors in the 5th-95th percentile (50th percentile would be the median).}
	\label{fig:error1dsfmvsours}
\end{figure}

\begin{figure}
	\setlength{\abovecaptionskip}{-0.3cm}
	\setlength{\belowcaptionskip}{-0.6cm}
	\vspace{-0.05cm}
	\begin{center}
		\includegraphics[width=1\linewidth, trim = 20mm 0mm 23mm 0mm, clip]{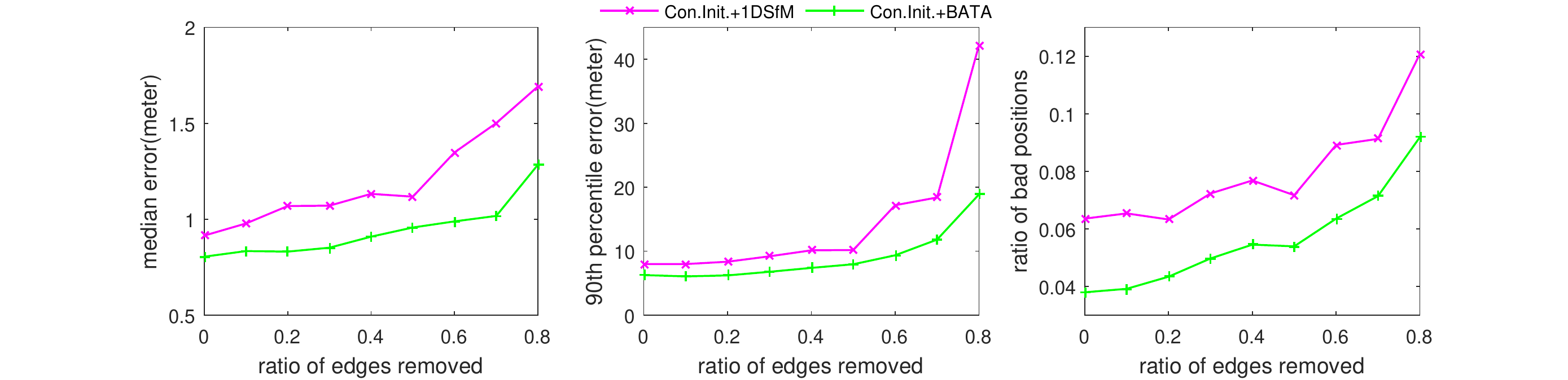}
	\end{center}
	\caption{Different error quantities plotted against the ratio of the observed edges removed.}
	\label{fig:dropping}
\end{figure}

We show these results in Tab.~\ref{tab:accuracycomparison}, together with those from four other state-of-the-art methods. Empirically We find BATA works well with a few different robust schemes, and here we only report the best results from Cauchy with $\alpha{=}0.1$; other results (e.g. Huber) are given in \emph{supp.} \emph{material}. Since Shapefit/kick \cite{goldstein2016shapefit} provides multiple results with different combinations of preprocessing strategies, we only cite the overall best one.  
The errors are given in terms of median distance error $\tilde{e}$ and mean distance error $\bar{e}$ between the estimated and the reference camera locations. 
The median distance error is used as a main measurement of quality since it better captures the accuracy of the overall shape of camera locations.  As can be seen, BATA obtains good accuracies even from random initialization. If bootstrapped by the convex method, the results generally improve. Compared to the na\"{\i}ve IRLS, the rotation-assisted IRLS generally improves the accuracies, especially in the case of random initialization. Also, ``\#iter" shows that it consistently reduces the number of iterations required for convergence.
We now compare BATA's result from the ``w Rot." case to those from the other four cited methods. The lowest median distance error is bolded for each scene. As can be seen, there is not one single best method for all the scenes. However, BATA gives the overall best performance in the sense that it achieves the lowest median distance errors $\tilde{e}$ in ten out of all the fourteen scenes. 
%We have also tested BATA with the preprocessing steps used in 1DSfM and LUD; no significant improvement is observed. 
We note that the method of \cite{cui2015global} generally achieves the lowest mean distance errors, which might be due to the local BA involved in their framework, making the estimation for those sparsely connected cameras less unstable. Also note that 1DSfM suffers from large mean errors. Next, we compare the ratio $r_1$ and $r_2$ computed from the LUD and Shapefit/kick result run on the raw view graph. As shown in Tab.~\ref{tab:realdataratio}, LUD generally returns a lower value of $r_1$ and $r_2$, indicating a squashing effect on the shape of the recovered camera locations. In view of the similarity of 1DSfM to BATA, we also test it on the raw view graph.
% (without involving feature points and outlier filtering as in \cite{wilson2014robust}). 
%We find the results are generally comparable to BATA in terms of median error. 
We find that although it can recover those well-conditioned cameras well, BATA generally recovers those sparsely connected camera positions more reliably even with na\"{\i}ve IRLS. To show this, we plot the distribution of errors in the NYC Library scene in Fig.~\ref{fig:error1dsfmvsours}. As highlighted by the ellipses, BATA achieves higher accuracies on those cameras with relatively large errors and we find these cameras often have a smaller number of links to others. This seems to indicate that BATA is more superior in handling sparsely connected cameras. To corroborate this, we have tested their performance under increasingly sparser view graph by manually removing the observed edges. We plot the median error,  the 90th percentile error and the ratio of cameras with large error (${>}20$m, termed as bad positions) against the ratio of edges removed in Fig.~\ref{fig:dropping}. As we can see, although the difference in median error is small, (b)\&(c) show that the two methods deviate from each other largely in their ability to localize those more ``problematic" cameras, especially when the view graph becomes increasingly sparser and more cameras become sparsely connected. We leave more results from other scenes to the \textit{supp. material}.
%For a more comprehensive comparison with the NL formulation in \cite{wilson2014robust}, we also test the latter's performance with the raw EG graph under different initializations, including random initialization, the result from "Con. Init" and "w Rot". The results are denoted as $\tilde{e}_r$, $\tilde{e}_c$ and $\tilde{e}_b$, respectively. We also compute $\sigma_n$ and $\sigma_b$, the standard deviations  over all trials for the errors in $\tilde{e}_r$ and $\tilde{e}$, the latter for the ``R.I. w/o R." case. The results are shown in Tab.~\ref{tab:nonlinear}. As can be seen, NL relies significantly on a good initialization to obtain accurate result; its results from random initialization yields higher standard deviations, implying that it suffers more from bad local minima. Note that with the same initialization from "Con. Init", BATA ("w/o Rot.") generally obtains higher accuracy than NL. 

Next, we feed the initial poses obtained from different methods into a final BA step, using the Ceres\cite{ceres-solver}-based pipeline in Theia \cite{theia-manual}. We present an example of the BA results on the NYC Library scene. We note that although different final BA schemes may give results of different qualities, adopting the same pipeline means that the results are only affected by, and thus indicate, the accuracy of the initial camera poses. We compare our result to that from the two magnitude-based methods. We use the implementation in \cite{theia-manual} to obtain the camera pose estimates from the full pipeline of LUD method. The results of Shapefit/kick were provided by the authors. As shown in Fig.~\ref{fig:pointcloud}, although all the methods can reconstruct the main building reasonably well, not all can reconstruct the detailed structures of the two sculptures nearby. As highlighted by the red ellipses, LUD fails to recover both of them, and Shapefit/kick is able to recover only the left one while the point cloud of the right one is somewhat blurred and shifted. BATA can recover both of them successfully and yields the most similar results to those of the Bundler.

Finally, we compare the running time of different methods. Ours are obtained on a normal PC with a 3.4 GHz Intel Core i7 CPU and 16GB memory. The results are given in Tab.~\ref{tab:timing}. $T_p$ denotes the general preprocessing time, which may include outlier filtering and pairwise translation re-estimation to improve the input quality \cite{wilson2014robust,ozyesil2015robust,goldstein2016shapefit}, or local depth estimation and local BA
%to simplify the problem to similarity averaging 
\cite{cui2015global}. $T_{ini}$ denotes the time for convex initialization in our method, $T_t$ the time for solving the translation averaging optimization,  and $T_\Sigma$ the total time. 
%Note that the time for rotation averaging and the final BA is not involved.  
As can be seen, since BATA directly processes the raw EG view graph and the sequence of sparse linear system of equations involved can be solved by highly efficient libraries, it is generally faster by several times.

%% file: Section/sec_conclusion.tex
\section{Conclusion}
In this paper, we advocate a return to angle-based objectives for translation averaging, proposing a simple yet effective bilinear formulation with a rotation-assisted IRLS scheme, achieving good empirical performance. This formulation also contributes to a better understanding of the behavior of the existing convex methods.

\noindent \textbf{Acknowledgements}. This work was partially supported by the Singapore PSF grant 1521200082 and the Singapore MOE Tier 1 grant R-252-000-637-112.

%% file: main.bbl
\begin{thebibliography}{10}\itemsep=-1pt

\bibitem{ceres-solver}
S.~Agarwal, K.~Mierle, and Others.
\newblock Ceres solver.
\newblock \url{http://ceres-solver.org}.

\bibitem{arie2012global}
M.~Arie-Nachimson, S.~Z. Kovalsky, I.~Kemelmacher-Shlizerman, A.~Singer, and
  R.~Basri.
\newblock Global motion estimation from point matches.
\newblock In {\em 3DIMPVT}, 2012.

\bibitem{arrigoni2014robust}
F.~Arrigoni, L.~Magri, B.~Rossi, P.~Fragneto, and A.~Fusiello.
\newblock Robust absolute rotation estimation via low-rank and sparse matrix
  decomposition.
\newblock In {\em 3DV}, 2014.

\bibitem{chatterjee2013efficient}
A.~Chatterjee and V.~Madhav~Govindu.
\newblock Efficient and robust large-scale rotation averaging.
\newblock In {\em ICCV}, 2013.

\bibitem{crandall2011discrete}
D.~Crandall, A.~Owens, N.~Snavely, and D.~Huttenlocher.
\newblock Discrete-continuous optimization for large-scale structure from
  motion.
\newblock In {\em CVPR}, 2011.

\bibitem{cui2015global}
Z.~Cui and P.~Tan.
\newblock Global structure-from-motion by similarity averaging.
\newblock In {\em ICCV}, 2015.

\bibitem{enqvist2011non}
O.~Enqvist, F.~Kahl, and C.~Olsson.
\newblock Non-sequential structure from motion.
\newblock In {\em ICCV Workshops}, 2011.

\bibitem{fredriksson2012simultaneous}
J.~Fredriksson and C.~Olsson.
\newblock Simultaneous multiple rotation averaging using lagrangian duality.
\newblock In {\em ACCV}, 2012.

\bibitem{goldstein2016shapefit}
T.~Goldstein, P.~Hand, C.~Lee, V.~Voroninski, and S.~Soatto.
\newblock Shapefit and shapekick for robust, scalable structure from motion.
\newblock In {\em ECCV}, 2016.

\bibitem{govindu2001combining}
V.~M. Govindu.
\newblock Combining two-view constraints for motion estimation.
\newblock In {\em CVPR}, 2001.

\bibitem{hartley2013rotation}
R.~Hartley, J.~Trumpf, Y.~Dai, and H.~Li.
\newblock Rotation averaging.
\newblock {\em International journal of computer vision}, 103(3):267--305,
  2013.

\bibitem{hartley2003multiple}
R.~Hartley and A.~Zisserman.
\newblock {\em Multiple view geometry in computer vision}.
\newblock Cambridge university press, 2003.

\bibitem{jiang2013global}
N.~Jiang, Z.~Cui, and P.~Tan.
\newblock A global linear method for camera pose registration.
\newblock In {\em ICCV}, 2013.

\bibitem{kahl2005multiple}
F.~Kahl.
\newblock Multiple view geometry and the ${L}_\infty$-norm.
\newblock In {\em ICCV}, 2005.

\bibitem{martinec2007robust}
D.~Martinec and T.~Pajdla.
\newblock Robust rotation and translation estimation in multiview
  reconstruction.
\newblock In {\em CVPR}, 2007.

\bibitem{moulon2013global}
P.~Moulon, P.~Monasse, and R.~Marlet.
\newblock Global fusion of relative motions for robust, accurate and scalable
  structure from motion.
\newblock In {\em ICCV}, 2013.

\bibitem{moulon2016openmvg}
P.~Moulon, P.~Monasse, R.~Perrot, and R.~Marlet.
\newblock Openmvg: Open multiple view geometry.
\newblock In {\em International Workshop on Reproducible Research in Pattern
  Recognition}, pages 60--74. Springer, 2016.

\bibitem{nie2010efficient}
F.~Nie, H.~Huang, X.~Cai, and C.~H. Ding.
\newblock Efficient and robust feature selection via joint $l_{2,1}$-norms
  minimization.
\newblock In {\em NIPS}, 2010.

\bibitem{olsson2011stable}
C.~Olsson and O.~Enqvist.
\newblock Stable structure from motion for unordered image collections.
\newblock {\em Image Analysis}, pages 524--535, 2011.

\bibitem{ozyesil2015robust}
O.~Ozyesil and A.~Singer.
\newblock Robust camera location estimation by convex programming.
\newblock In {\em CVPR}, 2015.

\bibitem{schoenberger2016sfm}
J.~L. Sch\"{o}nberger and J.-M. Frahm.
\newblock Structure-from-motion revisited.
\newblock In {\em CVPR}, 2016.

\bibitem{Sengupta_2017_CVPR}
S.~Sengupta, T.~Amir, M.~Galun, T.~Goldstein, D.~W. Jacobs, A.~Singer, and
  R.~Basri.
\newblock A new rank constraint on multi-view fundamental matrices, and its
  application to camera location recovery.
\newblock In {\em CVPR}, 2017.

\bibitem{shen2016graph}
T.~Shen, S.~Zhu, T.~Fang, R.~Zhang, and L.~Quan.
\newblock Graph-based consistent matching for structure-from-motion.
\newblock In {\em ECCV}, 2016.

\bibitem{sim2006recovering}
K.~Sim and R.~Hartley.
\newblock Recovering camera motion using ${L}_\infty$ minimization.
\newblock In {\em CVPR}, 2006.

\bibitem{sinha2010multi}
S.~Sinha, D.~Steedly, and R.~Szeliski.
\newblock A multi-stage linear approach to structure from motion.
\newblock In {\em ECCV Workshops}, 2010.

\bibitem{snavely2006photo}
N.~Snavely, S.~M. Seitz, and R.~Szeliski.
\newblock Photo tourism: exploring photo collections in 3d.
\newblock In {\em ACM transactions on graphics (TOG)}, volume~25, pages
  835--846. ACM, 2006.

\bibitem{theia-manual}
C.~Sweeney.
\newblock Theia multiview geometry library: Tutorial \& reference.
\newblock \url{http://theia-sfm.org}.

\bibitem{sweeney2015optimizing}
C.~Sweeney, T.~Sattler, T.~Hollerer, M.~Turk, and M.~Pollefeys.
\newblock Optimizing the viewing graph for structure-from-motion.
\newblock In {\em ICCV}, 2015.

\bibitem{triggs1999bundle}
B.~Triggs, P.~F. McLauchlan, R.~I. Hartley, and A.~W. Fitzgibbon.
\newblock Bundle adjustment—a modern synthesis.
\newblock In {\em International workshop on vision algorithms}, pages 298--372.
  Springer, 1999.

\bibitem{tron2014distributed}
R.~Tron and R.~Vidal.
\newblock Distributed 3-d localization of camera sensor networks from 2-d image
  measurements.
\newblock {\em IEEE Transactions on Automatic Control}, 59(12):3325--3340,
  2014.

\bibitem{tron2016survey}
R.~Tron, X.~Zhou, and K.~Daniilidis.
\newblock A survey on rotation optimization in structure from motion.
\newblock In {\em CVPR Workshops}, 2016.

\bibitem{wilson2014robust}
K.~Wilson and N.~Snavely.
\newblock Robust global translations with 1dsfm.
\newblock In {\em ECCV}, 2014.

\bibitem{wu2011visualsfm}
C.~Wu et~al.
\newblock Visualsfm: A visual structure from motion system.
\newblock 2011.

\bibitem{yuan2006model}
M.~Yuan and Y.~Lin.
\newblock Model selection and estimation in regression with grouped variables.
\newblock {\em Journal of the Royal Statistical Society: Series B (Statistical
  Methodology)}, 68(1):49--67, 2006.

\bibitem{zach2010disambiguating}
C.~Zach, M.~Klopschitz, and M.~Pollefeys.
\newblock Disambiguating visual relations using loop constraints.
\newblock In {\em CVPR}, 2010.

\end{thebibliography}
